\documentclass[nohyperref]{article}

\usepackage{microtype}
\usepackage{graphicx}
\usepackage{subfigure}

\usepackage{booktabs} %
\usepackage{natbib}

\usepackage[colorlinks,citecolor=DarkGreen,linkcolor=FireBrick,urlcolor=FireBrick]{hyperref}

\usepackage{graphicx}
\usepackage{tikz}
\usepackage{tikz-cd}
\usepackage{times}
\usepackage{courier}
\usepackage{bm}
\usepackage{physics}
\usepackage{xcolor}
\usepackage{natbib}
\usepackage{mdframed}
\usepackage{nicefrac}
\usepackage{booktabs}
\usepackage{blindtext}
\usepackage{braket}
\usepackage{amsmath}
\usepackage{amssymb}
\usepackage{mathtools}
\usepackage{titletoc}
\usepackage{enumitem}

\usepackage{graphicx}
\usepackage{tikz}
\usepackage{tikz-cd}
\usepackage{times}
\usepackage{courier}
\usepackage{bm}
\usepackage{physics}
\usepackage{xcolor}
\usepackage{natbib}
\usepackage{mdframed}
\usepackage{nicefrac}
\usepackage{booktabs}
\usepackage{lipsum}
\usepackage{titlesec}
\usepackage{wrapfig,lipsum,booktabs}
\usepackage{authblk}
\usepackage{blindtext}
\usepackage{titletoc}
\usepackage{titletoc}
\usepackage{setspace}

\usepackage{authblk}
\definecolor{DarkGreen}{rgb}{0,0.40,0}
\definecolor{FireBrick}{rgb}{0.698,0.133,0.133}

\usepackage[capitalize,noabbrev]{cleveref}

\usepackage{CJK}

\newenvironment{claim}{  \begin{mdframed}[linecolor=black!0,backgroundcolor=black!10]\noindent%
		\ignorespaces}{\end{mdframed}}

\renewenvironment{figure}[1][]{
  \begin{originalfigure}[#1]
    \begin{mdframed}[linecolor=black!0,backgroundcolor=black!1]
}{
    \end{mdframed}
  \end{originalfigure}
}

\usepackage{array}
\usepackage{makecell}

\def\half{{\frac{1}{2}}}

\newcommand{\bea}{\begin{eqnarray}}
\newcommand{\eea}{\end{eqnarray}}

\usepackage{slashed}

\def\({\left(}
\def\){\right)}
\def\[{\left[}
\def\]{\right]}

\usepackage{dashbox}
\usepackage{xcolor}
\usepackage{colortbl}
\usepackage{arydshln}
\definecolor{lightyellow}{rgb}{1.0, 0.95, 0.7}
\definecolor{Blue}{rgb}{0, 0, 0.8}
\definecolor{blue}{rgb}{0,0,1}
\definecolor{darkgreen}{rgb}{0,0.40,0}
\definecolor{firebrick}{rgb}{0.698,0.133,0.133}

\definecolor{colorA}{rgb}{1,0,0}
\definecolor{colorB}{rgb}{0,0.3,1}
\definecolor{colorC}{rgb}{0.9,0.8,0.2}
\definecolor{colorD}{rgb}{0,0.65,0}
\definecolor{lesslightgray}{rgb}{0.5,0.5,0.5}
\definecolor{light-gray}{gray}{0.95}

\let\tilde\widetilde

\newcommand{\calF}{\mathcal{F}}

\newcommand{\calO}{\mathcal{O}}

\newcommand{\calS}{\mathcal{S}}
\newcommand{\calT}{\mathcal{T}}

\newcommand{\calV}{\mathcal{V}}

\newcommand{\bA}{\mathbf{A}}

\newcommand{\bD}{\mathbf{D}}

\newcommand{\bU}{\mathbf{U}}
\newcommand{\bV}{\mathbf{V}}

\newcommand{\bX}{\mathbf{X}}
\newcommand{\bY}{\mathbf{Y}}
\newcommand{\bZ}{\mathbf{Z}}
\newcommand{\ba}{\mathbf{a}}
\newcommand{\bb}{\mathbf{b}}

\newcommand{\bu}{\mathbf{u}}

\newcommand{\bv}{\mathbf{v}}

\newcommand{\bx}{\mathbf{x}}
\newcommand{\by}{\mathbf{y}}
\newcommand{\bz}{\mathbf{z}}
\newcommand{\bxi}{{\bm{\xi}}}
\newcommand{\diag}{\mathop{\rm{diag}}}
\newcommand{\Max}{\mathop{\rm Max}}
\newcommand{\Min}{\mathop{\rm Min}}

\newcommand{\Softmax}{\mathop{\rm{Softmax}}}

\newcommand{\lse}{\mathop{\rm{lse}}}

\newcommand{\sT}{ \mathsf{T} }

\newcommand{\sumM}{\sum_{\mu=1}^M}

\def\R{\mathbb{R}}

\let\cite\citep 

\usepackage{amsthm}
\makeatletter
\def\th@remark{%
  \thm@headfont{\bfseries}%
  \normalfont %
  \thm@preskip\topsep \divide\thm@preskip\tw@
  \thm@postskip\thm@preskip
}
\makeatother
\usepackage[many]{tcolorbox}
\theoremstyle{definition}
\newtheorem{theorem}{Theorem}[section]
\tcolorboxenvironment{theorem}{
  breakable,
  colback=black!10,
  colframe=white,%
  width=\dimexpr\linewidth+10pt\relax,%
  enlarge left by=-5pt,%
  enlarge right by=-5pt,%
  boxsep=5pt,%
  boxrule=0pt,
  left=0pt,right=0pt,top=0pt,bottom=0pt,
  sharp corners,
  before skip=\topsep,
  after skip=\topsep
}
\newtheorem{lemma}{Lemma}[section]
\tcolorboxenvironment{lemma}{
  breakable,
  colback=black!10,
  colframe=white,%
  width=\dimexpr\linewidth+10pt\relax,%
  enlarge left by=-5pt,%
  enlarge right by=-5pt,%
  boxsep=5pt,%
  boxrule=0pt,
  left=0pt,right=0pt,top=0pt,bottom=0pt,
  sharp corners,
  before skip=\topsep,
  after skip=\topsep
}
\newtheorem{corollary}{Corollary}[theorem]
\tcolorboxenvironment{corollary}{
  breakable,
  colback=black!10,
  colframe=white,%
  width=\dimexpr\linewidth+10pt\relax,%
  enlarge left by=-5pt,%
  enlarge right by=-5pt,%
  boxsep=5pt,%
  boxrule=0pt,
  left=0pt,right=0pt,top=0pt,bottom=0pt,
  sharp corners,
  before skip=\topsep,
  after skip=\topsep
}

\theoremstyle{definition}
\newtheorem{definition}{Definition}[section]
\tcolorboxenvironment{definition}{
  breakable,
  colback=black!10,
  colframe=white,%
  width=\dimexpr\linewidth+10pt\relax,%
  enlarge left by=-5pt,%
  enlarge right by=-5pt,%
  boxsep=5pt,%
  boxrule=0pt,
  left=0pt,right=0pt,top=0pt,bottom=0pt,
  sharp corners,
  before skip=\topsep,
  after skip=\topsep
}
\theoremstyle{remark}
\newtheorem{remark}{Remark}[section]

\tcolorboxenvironment{assumption}{
  breakable,
  colback=black!10,
  colframe=white,%
  width=\dimexpr\linewidth+10pt\relax,%
  enlarge left by=-5pt,%
  enlarge right by=-5pt,%
  boxsep=5pt,%
  boxrule=0pt,
  left=0pt,right=0pt,top=0pt,bottom=0pt,
  sharp corners,
  before skip=\topsep,
  after skip=\topsep
}
\newtheorem{problem}{Problem}
\tcolorboxenvironment{problem}{
  breakable,
  colback=black!10,
  colframe=white,%
  width=\dimexpr\linewidth+10pt\relax,%
  enlarge left by=-5pt,%
  enlarge right by=-5pt,%
  boxsep=5pt,%
  boxrule=0pt,
  left=0pt,right=0pt,top=0pt,bottom=0pt,
  sharp corners,
  before skip=\topsep,
  after skip=\topsep
}
\newtheorem{question}{Question}
\tcolorboxenvironment{question}{
  breakable,
  colback=black!10,
  colframe=white,%
  width=\dimexpr\linewidth+10pt\relax,%
  enlarge left by=-5pt,%
  enlarge right by=-5pt,%
  boxsep=5pt,%
  boxrule=0pt,
  left=0pt,right=0pt,top=0pt,bottom=0pt,
  sharp corners,
  before skip=\topsep,
  after skip=\topsep
}
\newtheorem{hypothesis}{Hypothesis}
\tcolorboxenvironment{hypothesis}{
  breakable,
  colback=black!10,
  colframe=white,%
  width=\dimexpr\linewidth+10pt\relax,%
  enlarge left by=-5pt,%
  enlarge right by=-5pt,%
  boxsep=5pt,%
  boxrule=0pt,
  left=0pt,right=0pt,top=0pt,bottom=0pt,
  sharp corners,
  before skip=\topsep,
  after skip=\topsep
}
\crefname{hypothesis}{Hypothesis}{Hypothesises}

\crefname{theorem}{Theorem}{Theorems}
\crefname{proposition}{Proposition}{Propositions}
\crefname{lemma}{Lemma}{Lemmas}
\crefname{corollary}{Corollary}{Corollaries}
\crefname{definition}{Definition}{Definitions}
\crefname{assumption}{Assumption}{Assumptions}
\crefname{remark}{Remark}{Remarks}
\crefname{problem}{Problem}{Problems}
\crefname{question}{Question}{Questions}
\crefname{property}{Property}{property}

\numberwithin{equation}{section}
\numberwithin{theorem}{section}
\numberwithin{proposition}{section}
\numberwithin{definition}{section}
\numberwithin{lemma}{section}
\numberwithin{assumption}{section}
\numberwithin{remark}{section}

\usepackage{lipsum}

\newcommand*{\annot}[1]{\tag*{\footnotesize{\textcolor{black!50}{\big(#1\big)}}}}

\makeatletter
\let\save@mathaccent\mathaccent
\newcommand*\if@single[3]{%
    \setbox0\hbox{${\mathaccent"0362{#1}}^H$}%
    \setbox2\hbox{${\mathaccent"0362{\kern0pt#1}}^H$}%
    \ifdim\ht0=\ht2 #3\else #2\fi
}
\newcommand*\rel@kern[1]{\kern#1\dimexpr\macc@kerna}
\newcommand*\widebar[1]{\@ifnextchar^{{\wide@bar{#1}{0}}}{\wide@bar{#1}{1}}}
\newcommand*\wide@bar[2]{\if@single{#1}{\wide@bar@{#1}{#2}{1}}{\wide@bar@{#1}{#2}{2}}}
\newcommand*\wide@bar@[3]{%
    \begingroup
    \def\mathaccent##1##2{%
        \let\mathaccent\save@mathaccent
        \if#32 \let\macc@nucleus\first@char \fi
        \setbox\z@\hbox{$\macc@style{\macc@nucleus}_{}$}%
        \setbox\tw@\hbox{$\macc@style{\macc@nucleus}{}_{}$}%
        \dimen@\wd\tw@
        \advance\dimen@-\wd\z@
        \divide\dimen@ 3
        \@tempdima\wd\tw@
        \advance\@tempdima-\scriptspace
        \divide\@tempdima 10
        \advance\dimen@-\@tempdima
        \ifdim\dimen@>\z@ \dimen@0pt\fi
        \rel@kern{0.6}\kern-\dimen@
        \if#31
        \overline{\rel@kern{-0.6}\kern\dimen@\macc@nucleus\rel@kern{0.4}\kern\dimen@}%
        \advance\dimen@0.4\dimexpr\macc@kerna
        \let\final@kern#2%
        \ifdim\dimen@<\z@ \let\final@kern1\fi
        \if\final@kern1 \kern-\dimen@\fi
        \else
        \overline{\rel@kern{-0.6}\kern\dimen@#1}%
        \fi
    }%
    \macc@depth\@ne
    \let\math@bgroup\@empty \let\math@egroup\macc@set@skewchar
    \mathsurround\z@ \frozen@everymath{\mathgroup\macc@group\relax}%
    \macc@set@skewchar\relax
    \let\mathaccentV\macc@nested@a
    \if#31
    \macc@nested@a\relax111{#1}%
    \else
    \def\gobble@till@marker##1\endmarker{}%
    \futurelet\first@char\gobble@till@marker#1\endmarker
    \ifcat\noexpand\first@char A\else
    \def\first@char{}%
    \fi
    \macc@nested@a\relax111{\first@char}%
    \fi
    \endgroup
    }
\makeatother
\let\bar\widebar

\makeatletter
\newcommand*{\redefinesymbolwitharg}[1]{%
  \expandafter\let\csname ltx#1\expandafter\endcsname\csname #1\endcsname
  \@namedef{#1}{\@ifnextchar{^}{\@nameuse{#1@}}{\@nameuse{#1@}^{}}}%
  \expandafter\def\csname #1@\endcsname^##1##2{%
     \csname ltx#1\endcsname\ifx!##1!\else^{##1}\fi\mathopen{}\mathclose\bgroup\left(##2\aftergroup\egroup\right)
     }%
}
\makeatother
\redefinesymbolwitharg{sin}
\redefinesymbolwitharg{cos}

\usepackage{url}
\usepackage{amsmath}
\usepackage{amssymb}
\usepackage{mathtools}
\usepackage{amsthm}
\usepackage{amsmath}
\usepackage{dsfont}

\usepackage[accepted]{icml2024}

\usepackage[textsize=tiny]{todonotes}

\icmltitlerunning{On Computational Limits of Modern Hopfield Models: 
A Fine-Grained Complexity Analysis}

\setlist[itemize]{leftmargin=1em, before=\vspace{-0.5em}, after=\vspace{-0.5em}, itemsep=0.1em}
\setlist[enumerate]{leftmargin=1.2em, before=\vspace{-0.5em}, after=\vspace{-0.5em}, itemsep=0.1em}

\begin{document}

\twocolumn[
\icmltitle{On Computational Limits of Modern Hopfield Models:\\
A Fine-Grained Complexity Analysis
}

\icmlsetsymbol{equal}{*}

\begin{icmlauthorlist}
\icmlauthor{Jerry Yao-Chieh Hu}{equal,yyy}
\icmlauthor{Thomas Lin}{equal,comp}
\icmlauthor{Zhao Song}{adobe}
\icmlauthor{Han Liu}{sch}

\end{icmlauthorlist}

\icmlaffiliation{yyy}{Department of Computer Science, Northwestern University, Evanston, IL, USA}
\icmlaffiliation{comp}{Department of Physics, National Taiwan University, Taipei, Taiwan }
\icmlaffiliation{adobe}{Adobe Research, Seattle, WA, USA}
\icmlaffiliation{sch}{Department of Statistics and Data Science, Northwestern University, Evanston, IL, USA}

\icmlcorrespondingauthor{Jerry Yao-Chieh Hu}{\href{mailto:jhu@u.northwestern.edu}{jhu@u.northwestern.edu}}
\icmlcorrespondingauthor{Thomas Lin}{\href{mailto:b12202026@ntu.edu.tw}{b12202026@ntu.edu.tw}}
\icmlcorrespondingauthor{Zhao Song}{\href{mailto:zsong@adobe.com}{zsong@adobe.com}}
\icmlcorrespondingauthor{Han Liu}{\href{mailto:hanliu@northwestern.edu}{hanliu@northwestern.edu}}

\icmlkeywords{Machine Learning, ICML}

\vskip 0.3in
]

\printAffiliationsAndNotice{\icmlEqualContribution} %

\titlespacing*{\section}{0pt}{0pt}{0pt}
\titlespacing*{\subsection}{0pt}{0pt}{0pt}
\titlespacing*{\subsubsection}{0pt}{0pt}{0pt}
\begin{abstract}
We investigate the computational limits of the memory retrieval dynamics of modern Hopfield models from the fine-grained complexity analysis.
Our key contribution is the characterization of a phase transition behavior in the efficiency of all possible modern Hopfield models based on the norm of patterns.
Specifically, we establish an upper bound criterion for the norm of input query patterns and memory patterns.
Only below this criterion, sub-quadratic (efficient) variants of the modern Hopfield model exist, assuming the Strong Exponential Time Hypothesis (SETH).
To showcase our theory, we provide a formal example of efficient constructions of modern Hopfield models using low-rank approximation when the efficient criterion holds.
This includes a derivation of 
a lower bound on the computational time, scaling linearly with $\max\{$\# of stored memory patterns, length of input query sequence$\}$.
In addition, we prove its
memory retrieval error bound and exponential memory capacity.

\end{abstract}

\section{Introduction}
\label{sec:intro}
We investigate the computational limits of modern Hopfield models \cite{wu2024uniform,wu2023stanhop,hu2024outlier,hu2024nonparametric,hu2023SparseHopfield,ramsauer2020hopfield} from a fine-grained complexity analysis, and characterize a norm-based phase transition for all possible efficient modern Hopfield model.
This analysis holds practical significance. 
Modern Hopfield models are a type of associative memory model compatible with deep learning.
More precisely, their deep learning derivatives offer robust alternatives to attention mechanisms in various transformer- and Hopfield-based methods \cite{hofmann2024energy,xu2024bishop,wu2024uniform,wu2023stanhop,hu2024outlier,schimunek2023contextenriched,furst2022cloob,paischer2022history,seidl2022improving,widrich2020modern}.
However, these models currently lack efficient implementations for large-scale applications  \citep[Section~C.2]{hu2023SparseHopfield}. 
This issue becomes more relevant with the rise of Large Foundation Models  \cite{bommasani2021opportunities}, where expansive attention-based architectures, pre-trained on vast datasets, are pivotal across multiple scientific fields, including natural language processing \cite{brown2020language,floridi2020gpt}, financial analytics  \cite{wu2023bloomberggpt}, genomic research \cite{zhou2024dnabert,zhou2023dnabert,ji2021dnabert}, medical science \cite{thirunavukarasu2023large,singhal2023large,moor2023foundation} and more. 
This work makes a timely theoretical analysis of their computational limits, aimed at advancing (Hopfield-based) large foundation models.

Let $\bx \in \mathbb{R}^d$ be the input query pattern. The memory patterns are stored in a matrix $\bm{\Xi} = [\bxi_1, \cdots, \bxi_M] \in \mathbb{R}^{d \times M}$. 
Hopfield models are energy-based associative memory models.
These models store memory patterns $\bm{\Xi}$ on the local minima of their energy landscapes, i.e. energy functions $E$.
For any input query $\bx$, they retrieve its closest memory pattern through some energy minimization algorithms, i.e. retrieval dynamics $\calT$, initialized at $\bx$.

\citet{ramsauer2020hopfield} propose the Modern Hopfield Model with a specific set of energy function $E$ and memory retrieval dynamics $\calT$, and integrate it into deep learning architectures via its connection with the transformer attention \cite{vaswani2017attention}, offering enhanced performance, and theoretically guaranteed exponential memory capacity.
Specifically, they introduce the energy function:
\begin{equation}
\label{eqn:MHM}
    E(\bx) = -\lse(\beta,\bm{\Xi}^\sT \bx) + \frac{1}{2} \langle \bx,\bx \rangle ,
\end{equation}
where the retrieval dynamics is given by 
\bea
\label{eqn:retrival_dyn}
\bx^{\text{new}}=\calT_{\text{Dense}}(\bx) = \bm{\Xi} \cdot \Softmax(\beta \bm{\Xi}^\sT \bx).
\eea
The function $\lse\(\beta,\bz\)\coloneqq \log\(\sumM \exp{\beta z_\mu}\)/\beta$ is the log-sum-exponential for any given vector $\bz\in\R^M$ and $\beta>0$.
Let $\bX=\[\bx_1,\ldots,\bx_L\]\in \R^{d\times L}$ be a sequence of input queries, such that \eqref{eqn:retrival_dyn} becomes 
$\bZ\coloneqq
\[\bx_1^{\text{new}},\ldots,\bx_L^{\text{new}}\]
=\calT_{\text{Dense}}(\bX)$, and hence
\begin{align*}
\calT_{\text{Dense}}(\bX)
= \overbrace{\bm{\Xi}}^{\in\R^{d\times M}} \cdot \Softmax(\beta \overbrace{\underbrace{\bm{\Xi}^\sT}_{\in\R^{M\times d}} \underbrace{\bX}_{\in\R^{d\times L}}}^{\in\R^{M\times L}})\in\R^{d\times L},
\end{align*}
where the $\Softmax(\cdot)$ applies column-wise normalization\footnote{Many existing works denote $\bZ$ by $\bX^{\text{new}}.$}.
Here we assume $d=L^{o(1)}$, i.e.,  the growth rate of this function is sub-polynomial concerning $L$.

To motivate the study of possible efficient implementations, we make the following observation on \eqref{eqn:MHM}:
\begin{claim}
    The bottleneck of Hopfield-based methods is the time to perform matrix multiplication in memory retrieval: 
    $\calO(dML)$.
    Namely, \eqref{eqn:MHM} is  inefficient with  $M=\Omega(e^{d})$ (large memory set) and $L=\Omega(e^{d})$ (long query sequences).
\end{claim}
Explicitly, if the associative space is $d$-dimensional, this necessitates $d$ multiplication operations for the inner products of $\{\mathbf{x}\}$ and $\{\boldsymbol{\xi}\}$. 
Consequently, the complexity of computing a dot product is $\mathcal{O}(d)$. Each pattern in $\mathbf{Z}$ must associate with every pattern in $\boldsymbol{\Xi}$. 
Therefore, the time complexity for sequences of length $L$ and $M$ with a pattern dimension of $d$ is $\mathcal{O}(dML)$.
In this regard, this work aims to characterize the fundamental limits on improving $\calO(dML)$.
Specifically, we ask the following questions:
\begin{question}
\label{prob:1}
    Is it possible to improve the time complexity $\calO(dML)$ with a controllable approximation error?
\end{question}
\begin{question}
\label{prob:2}
    More aggressively, is it possible to perform memory retrieval computations in almost linear time $L^{1+o(1)}$ or $M^{1+o(1)}$ or $(L+M)^{1+o(1)}$? 
\end{question}
To address these questions, we explore approximate retrieval computations with precision guarantees.
We aim to find a surrogate $\calT_{\text{approx.}}$ (also denoted as $\Tilde{\calT}_{\text{Dense}}$) for $\calT_{\text{Dense}}$ such that
\begin{align*}
\norm{\calT_{\text{approx.}}-\calT_{\text{Dense}}}_{\text{max}}\le \delta_{\text{approx.}},
\end{align*}
for some $\delta_{\text{approx.}}>0$, where $\norm{\bA}_{\text{max}}\coloneqq \max_{i,j}\abs{a_{ij}}$.

To be concrete, we study the following approximation problem with the realistic setting $\delta_{\text{approx.}}=1/\text{poly}(L)$.

\begin{problem}[Approximate Modern Hopfield Memory Retrieval Dynamics  $\mathtt{AHop}(d,M,L,\beta,B,\delta_H)$]
\label{prob:AHop}
    Let $\delta_H > 0$. 
    Given $\bm\Xi \in \R^{d\times M}$ and $\bX \in \R^{d\times L}$ such that $\norm{\bm\Xi}_{\text{max}} \leq B$ and $ \norm{\bX}_{\text{max}} \leq B$.
    We aim to study an approximation problem $\mathtt{AHop}(d,M,L,\beta,B,\delta_H)$, that approximates $\bZ$ with a matrix $\Tilde{\bZ}\coloneqq\Tilde{\calT}_{\text{Dense}}(\bX)$ such that
    \begin{align*}
        \norm{\Tilde{\bZ} - \bm\Xi\bD^{-1}\bA}_{\text{max}} \leq \delta_H,
    \end{align*}
    where $\bm\Xi\bD^{-1}\bA=\bZ$ with
    \begin{align*}
        \bA = \exp{\beta\bm\Xi^\sT\bX}, \quad \bD=\diag(\bA\textbf{1}_M).
    \end{align*}
    
\end{problem}

In this work, we aim to investigate the computational limits and potential efficient algorithms of $\mathtt{AHop}$. 

\paragraph{Contributions.}
Our contributions are threefold:
\begin{itemize}
    \item \textbf{Computational Limits.} 
    We answer \cref{prob:1} by identifying a phase transition behavior on the norm of query and memory patterns assuming the Strong Exponential Time Hypothesis (SETH). 
    Explicitly, 
    let $\tau = \max\left\{M,L\right\}$ be the upper bound of the patterns' lengths.
    We prove an upper bound criterion $B^\star = \Theta(\sqrt{\log \tau})$ for $\norm{\bm\Xi}_{\text{max}}$ and $\norm{\bX}_{\text{max}}$ such that, only below which, solving $\mathtt{AHop}$ in $\tau^{2-\Omega(1)}$ (sub-quadratic) time is possible.

    \item \textbf{Efficient Model.} We answer \cref{prob:2} by providing an efficient algorithm for $\mathtt{AHop}$ based on low-rank approximation: an almost linear time modern Hopfield model. 
    Explicitly, 
    we prove that the algorithm, under realistic settings, performs the computation in almost linear time $\tau^{1+o(1)}$.
    
    \item 
    \textbf{Exponential Memory Capacity.} 
    Focusing on the almost-linear-time modern Hopfield model, we derive its retrieval error bound and show that this model achieves almost-linear-time efficiency while maintaining the exponential memory capacity characteristic of modern Hopfield models.
\end{itemize}

\subsection*{Background and Related Works}

\paragraph{Modern Hopfield Models for Deep Learning.}
Classical Hopfield models \cite{hopfield1984neurons, hopfield1982neural, krotov2016dense} emulate human brain associative memory by focusing on storing and retrieving memory patterns.
In machine learning community, a noticeable interest in these models arises from (i) improved memory storage capacities (from linear to polynomial \cite{krotov2016dense}, to exponential \cite{demircigil2017model} and to kernelized \cite{wu2024uniform}), (ii) novel architectures \cite{hoover2023energy,seidl2022improving,furst2022cloob}, and (iii) their biological plausibility \cite{kozachkov2022building,krotov2021large}.
Notably, the modern Hopfield models \cite{hu2024outlier,hu2024nonparametric,hu2023SparseHopfield,wu2024uniform,wu2023stanhop,burns2023simplicial,hopfieldblog2021,ramsauer2020hopfield} offer fast convergence and expanded memory capacity. 
Importantly, they serve as advanced extensions of attention mechanisms to Transformer architecture.
They have extensive applications in diverse fields like tabular learning \cite{xu2024bishop}, drug discovery \cite{schimunek2023contextenriched}, immunology \cite{widrich2020modern}, time series forecasting \cite{wu2023stanhop,auer2023conformal}, reinforcement learning \cite{paischer2022history}, and large foundation models \cite{hu2024outlier,furst2022cloob}.

\paragraph{Theory of Modern Hopfield Models.} 
Besides empirical successes, 
Modern Hopfield Models provide a model-based theoretical framework for analyzing transformer attention and Transformer architectures.
\cite{hu2023SparseHopfield} and \citet{wu2023stanhop} propose a unified framework to analyze and derive modern Hopfield models via entropic regularizers.
Significantly, their work presents sparse variants (sparse and generalized sparse models) and incorporates the standard modern Hopfield model \cite{ramsauer2020hopfield} as a particular example in their framework.
Yet, they also note that the modern Hopfield paradigm is incomplete and lacks efficient implementations or variants \citep[Section~E]{hu2023SparseHopfield}.
Extending this foundation, \citet{hu2024nonparametric} introduces a principled construction of efficient variants from the nonparametric perspective, including linear, top-K, and random feature modern Hopfield models. 
This study aims to refine this research direction towards efficient models. 
We believe that this study is critical in guiding future research toward a Hopfield-driven design paradigm, especially for large-scale models.

\paragraph{Fine-Grained Complexity.} 
Much of fine-grained complexity theory relies on hypotheses concerning the time complexity of three problems: Conjunctive Normal Form Satisfiability (CNFSAT), All-Pairs Shortest Paths (APSP), and 3-SUM \cite{williams2018some}. 
\citet{ip01} introduce the Strong Exponential Time Hypothesis (SETH) to address the complexity of CNF-SAT. 
SETH is a stronger form of the $\mathtt{P} \neq \mathtt{NP}$ conjecture, suggesting that our current best $\mathtt{SAT}$ algorithms are optimal.
It states as follows:
\begin{hypothesis}[SETH]
\label{hyp:seth}
For every $\epsilon > 0$, there is a positive integer $k \geq 3$ such that $k$-$\mathtt{SAT}$ on formulas with $n$ variables cannot be solved in $\calO(2^{(1-\epsilon )n})$ time, even by a randomized algorithm.
\end{hypothesis}
SETH is a popular conjecture for proving fine-grained lower bounds for a wide variety of algorithmic problems, such as $k$-Hitting Set and $k$-NAE-SAT \cite{cygan2016problems}. 
See \citet{williams2018some}  for a comprehensive review.
Along this line, 
we utilize the fine-grained reduction under SETH to analyze the computational limits.
In previous fine-grained reduction works, 
\citet{bis17} analyze the computational complexity for multiple Empirical Risk Minimization problems, such as kernel SVMs and kernel ridge. 
\citet{alman2020algorithms} study the applicability of efficient spectral graph theory on geometric graphs under SETH.
\citet{aggarwal2022optimal} focus on the computational limits of Batch Gaussian Kernel Density Estimation problems.
\citet{alman2023bypass} utilize the weight-data correlation in a tree data structure for fast neural network training. 
\citet{alman2023fast,as23_tensor} extend the previous work to transformer attention and introduce a tensor generalization. 
Compared to existing works,
this work is, to the best of our knowledge, the first analysis of computational limits for modern Hopfield (associative memory) models \cite{hu2024outlier,hu2024nonparametric,wu2024uniform,wu2023stanhop,hu2023SparseHopfield,ramsauer2020hopfield}.  
In addition,
it offers a more general characterization, encompassing computational analyses of self-attention \cite{as23_tensor,alman2023fast} and cross-attention as special cases.

\paragraph{Notations.} 
We denote (column) vectors by lower case bold letters, and matrices by upper case bold letters.
We write $\Braket{\ba,\bb}\coloneqq \ba^\sT \bb$ as the inner product for vectors $\ba,\bb$.
Let $\ba[i]$ denotes the $i$-th component of vector $\ba$.
The index set $\{1,\cdots,I\}$ is denoted by $[I]$, where $I\in\mathbb{N}_+$.
Let $\norm{\bA}_{\text{max}}\coloneqq\max_{i,j} \abs{\bA_{ij}}$ for any matrix $\bA$. 
We denote the memory patterns by $\bxi\in\R^d$ and the query pattern by $\bx\in\R^d$, and $\bm{\Xi}\coloneqq\[\bxi_1,\cdots,\bxi_M\]\in \R^{d\times M}$ as shorthand for stored memory patterns $\{\bxi_\mu\}_{\mu\in[M]}$.
We denote $\{\underline{\bm\tau}_{1}, \cdots, \underline{\bm\tau}_{d}\} \subset \R^{1\times n}$ for each row in the matrix $\bZ \in \R^{d\times n}$.

\section{Computational Limits}
\label{sec:computation}
In this section,
we characterize the computational limits of all possible efficient variants of modern Hopfield models, i.e. $\mathtt{AHop}$, via fine-grained reduction.
Our primary technique involves casting the $\mathtt{AHop}$ problem (\cref{prob:AHop}) as a subroutine in the Approximate Nearest Neighbor Search Problem and deducing the hardness through reduction.

\subsection{Background: Approximate Nearest Neighbor Search Problem }

Approximate Nearest Neighbor Search ($\mathtt{ANNS}$) problem \cite{indyk1998approximate, arya1998optimal, muja2014scalable, li2019approximate} shares the same objective with the $\mathtt{AHop}$ problem of identifying a pattern closely resembling a query pattern as a memory retrieval process. 
Furthermore, the $\mathtt{ANNS}$ problem, which is particularly useful in high-dimensional spaces, seeks an approximate nearest neighbor within acceptable bounds to avoid the prohibitive computational costs of finding the exact nearest neighbor \cite{indyk1998approximate, muja2014scalable}.
In this work,
we observe that $\mathtt{ANNS}$ aligns with the goal of memory retrieval to efficiently find and recall the most relevant memory pattern in response to a specific input query. 
In our context, this translates to approximating the largest entry of $\Softmax(\bm{\Xi}^\sT\bx)$ in \eqref{eqn:retrival_dyn} for each query $\bx$, while maintaining a bounded error.

In $\mathtt{ANNS}$, 
one is given as input $n$ vectors of dimension $d$, and an error parameter $\delta>0$, and the goal is to find a pair of vectors whose distance is at most $(1 + \delta)$ times the \textit{minimum} distance between any pair of the vectors. The straightforward algorithm for $\mathtt{ANNS}$ runs in quadratic time, and it is known that it is impossible to solve $\mathtt{ANNS}$ in truly sub-quadratic time assuming $\mathtt{SETH}$~\cite{r18}.

To be concrete, we state the $\mathtt{ANNS}$ problem considered in this work as follows. 
\begin{definition}[Approximate Nearest Neighbor Search $\mathtt{ANNS}$]
\label{def: ANN}
    Given $\delta > 0$, $(1+\delta)\text{-}\mathtt{ANNS}$ for sets $A, B \subset \{0,1\}^d$, with $|A|=|B|=n$ requires finding $\ba^* \in A, \bb^* \in B$ such that:
    \begin{equation}
    \label{eqn:ANN_opt}
        \norm{\ba^*-\bb^*}_2^2 \leq (1+\delta) \min_{\ba \in A, \bb \in B} \norm{\ba-\bb}_2^2.
    \end{equation}
\end{definition}

Next, we present the hardness results from \citet{r18} as an auxiliary lemma for later use. 
Specifically, \citet{r18} show that no sub-quadratic-time algorithms exist for the $\mathtt{ANNS}$.

\begin{lemma}[Hardness for $\mathtt{ANNS}$, Theorem 4.1 of \cite{r18}]
\label{lemma:2.6}
    Assuming \cref{hyp:seth}, for every $q>0$, there exist $\delta \in (0,0.1)$ and $C>0$ such that $(1+\delta)\text{-}\mathtt{ANNS}$ with dimension $d=C\log n$ requires $\Omega(n^{2-q})$ time.
\end{lemma}
\subsection{Fine-Grained Reduction for $\mathtt{AHop}$}

To study the computational limits,
our proof strategy involves connecting $\mathtt{AHop}$ to the hardness of $\mathtt{ANNS}$ (see \cref{lemma:2.6}) through a fine-grained reduction.
We do this by introducing a decision problem $\mathtt{Gap}\text{-}\mathtt{ANNS}$ as a $(1+\delta)$-gap reduction \cite{demaine2014algorithmic} of the $\mathtt{ANNS}$ optimization problem \eqref{eqn:ANN_opt}, making the analysis more tractable while maintaining the same level of hardness. 
To be more precise, if we prove $\mathtt{AHop}$ is a reduction of $\mathtt{Gap}\text{-}\mathtt{ANNS}$, we also prove $\mathtt{AHop}$ is a $(1+\delta)$-gap reduction of $\mathtt{ANNS}$.
We start with $\mathtt{Gap}\text{-}\mathtt{ANNS}$ in below.

\begin{definition}[Gap Approximate Nearest Neighbor Search $\mathtt{Gap}\text{-}\mathtt{ANNS}(d,n,t,\delta)$]
\label{def: gap-ann}
    Given two sets of $n$ input vectors $A = \{\ba_1, \ldots, \ba_n\} \subset \{0,1\}^d$ and $B = \{\bb_1, \ldots, \bb_n\} \subset \{0,1\}^d$, the $\mathtt{Gap}\text{-}\mathtt{ANNS}(d,n,t,\delta)$ problem requires, for each $i\in [n]$, distinguish between the following two cases:
    \begin{itemize}
        \item 
        \textbf{Case 1:} There exists at least one pair $(\ba_i, \bb_j) \in A \times B$ such that $\norm{\ba_i - \bb_j}_2^2 < t$.

        \item        
        \textbf{Case 2:}
        For all $\bb_j \in B$, it holds that $\norm{\ba_i - \bb_j}_2^2 \geq (1+\delta) \cdot t$.
    \end{itemize}

    An algorithm for $\mathtt{Gap}\text{-}\mathtt{ANNS}(d,n,t,\delta)$ with $\log (nd)$ time can binary search the answer of $\mathtt{ANNS}$ \cite{williams2018some}.
\end{definition}
Then, we show that $\mathtt{AHop}$ serves as a subroutine within $\mathtt{Gap}\text{-}\mathtt{ANNS}$, thereby establishing a connection between the computational complexities of both problems.
\begin{theorem}[Reduction from $\mathtt{ANNS}$ to $\mathtt{AHop}$]
\label{lemma:2.7}
    Consider $\mathtt{Gap}\text{-}\mathtt{ANNS}$ with two sets of $n$ input vectors, for every $q>0$, for any chosen constants $C, C_0>0$, 
    there exist $\delta \in (0,0.1)$ and constants $C_\alpha, C_\beta > 0$ such that: 
    $\mathtt{Gap}\text{-}\mathtt{ANNS}(d=C\log n, n, t=C_0\log n, \delta)$ requires $\calO(T + n^{2-q})$ time
    if $\mathtt{AHop}(2d, M=2n,L=2n,\beta=1/2d,B=C_\beta\sqrt{\log n},\delta_H = n^{-C_\alpha})$ requires time $T$.
\end{theorem}
\begin{proof}[Proof Sketch]

To solve $\mathtt{Gap}\text{-}\mathtt{ANNS}$,
we employ different approaches for two scenarios, either through 
\begin{itemize}
    \item 
    \textbf{Scenario 1:}
    a brute-force approach, or
    \item 
    \textbf{Scenario 2:}
    reducing $\mathtt{Gap}\text{-}\mathtt{ANNS}$ to an $\mathtt{AHop}$ problem, and translating $\mathtt{AHop}$'s solution to  $\mathtt{Gap}\text{-}\mathtt{ANNS}$'s solution (i.e. distinguish the 2 cases in \cref{def: gap-ann}).
\end{itemize}
The proof of \textbf{Scenario 1} employs a brute-force algorithm for $\mathtt{Gap}\text{-}\mathtt{ANNS}$. 
This algorithm iterates over vectors within a Hamming distance of $t$ from each input vector and checks for a match in the target set. 
It results in a manageable time complexity $\calO(n^{2-q})$.

The proof for \textbf{Scenario 2} adopts a complex strategy.
Initially, an $\mathtt{AHop}$ instance is formulated to encompass the $\mathtt{Gap}\text{-}\mathtt{ANNS}$ problem. This formulation involves selecting specific parameters to ensure that resolving the $\mathtt{AHop}$ problem concurrently addresses the $\mathtt{Gap}\text{-}\mathtt{ANNS}$ challenge. 
Next, we introduce $\Tilde{t}$, a threshold exceeding the $\mathtt{AHop}$ algorithm's error bound, to effectively bridge the conditions of the $\mathtt{Gap}\text{-}\mathtt{ANNS}$ problem with the compound inequality derived from $\mathtt{AHop}$.
Finally, by considering an illustrative set of input vectors under the premise of a uniform distribution, the method for resolving the $\mathtt{Gap}\text{-}\mathtt{ANNS}$ is elucidated with the established value of $\Tilde{t}$. This approach simplifies the decision-making process in solving $\mathtt{Gap}\text{-}\mathtt{ANNS}$. 
\end{proof}
\paragraph{Main Proof.}
Here is the main proof of \cref{lemma:2.7}.
\begin{proof}
\label{proof:hardness_reduction}
    Let $\{\ba_1,\cdots,\ba_n,\bb_1,\cdots,\bb_n\} \subseteq \{0,1\}^d$ denote the input vectors of $\mathtt{Gap}\text{-}\mathtt{ANNS}(d=C\log n, n, t, \delta)$. For any given $c$ satisfying
    \begin{equation}
        \label{eqn:c_condition}
        \left\{
        \begin{array}{l}
            c\left( \log C+1 \right) \leq 1-q \\
            0 < c\leq \frac{1}{2}C,
        \end{array}
        \right.
    \end{equation}
    we categorize two scenarios based on whether $t < c \log n$.

    \paragraph{Scenario 1:} $t<c\log n$.
    
        The brute-force algorithm is described below:
        \begin{enumerate}
            \item For each $i\in [n]$, iterate over all vectors $\bb^\prime \in \{0,1\}^d$ which have Hamming distance at most $t$ from $\ba_i$.
            \item Check whether $\bb^\prime \in \{\bb_1,\cdots,\bb_n\}$.
        \end{enumerate}
        Since $t < \frac{1}{2} C\log n < d$, there are ${d \choose t}$ choices for the vector $\bb^\prime$, so the algorithm takes $\calO(n \cdot {d \choose t})$ time. We know:
        \begin{align}
            n \cdot {d \choose t} &\leq n \cdot {C\log n \choose c\log n} \leq \left(e\frac{C}{c}\right)^{c\log n} 
            \leq n^{1+c\log(Ce)}. \nonumber
        \end{align}
        Therefore, if we choose constant $c$ satisfying \eqref{eqn:c_condition}, the algorithm requires $\calO(n^{2-q})$ time.

        \paragraph{Scenario 2:} $t\geq c \log n$. 
        
        \textbf{Scenario 2 - Part 1.}
        This part shows the associated $\mathtt{AHop}$ problem. 
        Our objective is to construct an instance of the $\mathtt{AHop}$ problem in such a way that solving it also addresses the $\mathtt{Gap}\text{-}\mathtt{ANNS}$ problem. 
        To this end, we configure the $\mathtt{AHop}(\tilde{d}, \tilde{n}, \tilde{n}, \beta, B, \delta_H)$ problem by selecting a specific set of parameters:
        \begin{align*}
            \Tilde{d} \coloneqq 2d, \quad \Tilde{n} \coloneqq 2n, \quad \beta \coloneqq 1/\Tilde{d},
        \end{align*}
        \begin{equation}
        \label{def: C_a/C_b}
            C_\beta > 2\sqrt{C/(C_0\delta)}, \quad C_\alpha > \frac{C_\beta^2}{4} (3 + C_0/C)+1, 
        \end{equation}
        \begin{equation}
        \label{def: B}
             \quad B \coloneqq C_\beta \sqrt{\log n}, \quad \delta_H \coloneqq n^{-C_\alpha},
        \end{equation}
        Note that $\delta_H$ is dependent on but not equal to $\delta$.

        We parametrize $\mathtt{AHop}$'s input, $\bm{\Xi}$ and $\bX$, with $\mathtt{Gap}\text{-}\mathtt{ANNS}$ input $\{\ba_1,\cdots,\ba_n,\bb_1,\cdots,\bb_n\}$: 
        \begin{equation}
            \bm\Xi \coloneqq B \cdot 
            \begin{bmatrix}
                \ba_1 & \ba_2 & \cdots & \ba_n & \textbf{0}_d &\textbf{0}_d & \cdots &\textbf{0}_d \\
                \textbf{1}_d & \textbf{1}_d & \cdots &\textbf{1}_d & \textbf{1}_d & \textbf{1}_d & \cdots &\textbf{1}_d
            \end{bmatrix} \in \R^{\Tilde{d}\times\Tilde{n}},  \nonumber
        \end{equation}
        \begin{equation}
            \bX \coloneqq B \cdot 
            \begin{bmatrix}
                \bb_1 & \bb_2 & \cdots & \bb_n & \textbf{0}_d &\textbf{0}_d & \cdots &\textbf{0}_d \\
                \textbf{0}_d & \textbf{0}_d & \cdots &\textbf{0}_d & \textbf{1}_d & \textbf{1}_d & \cdots &\textbf{1}_d
            \end{bmatrix} \in \R^{\Tilde{d}\times\Tilde{n}}. \nonumber
        \end{equation}
        
        By construction, we have
        $\norm{\bm\Xi}_{\text{max}} \leq B$ and $\norm{\bX}_{\text{max}} \leq B$.
        This follows that:
        \begin{equation}
            \norm{\beta \bm\Xi^\sT \bX }_{\text{max}} \leq \beta B^2 \Tilde{d} = B^2. \nonumber
        \end{equation}

        Consider the matrix $\bA \coloneqq \exp {\beta \bm\Xi^\sT \bX } \in \R^{\Tilde{n} \times \Tilde{n}}$:
        \begin{align}
        \label{eqn:A_matrix_compact}
            \bA=
            \begin{bmatrix}
                \bA_1 & \bA_2\\
                \bA_3 & \bA_4
            \end{bmatrix}, 
        \end{align} 
        where $\bA_1,\bA_2,\bA_3,\bA_4\in\R^{n\times n}$: 
        \begin{align*}
            &\bA_1\coloneqq \left[\exp{\beta B^2 \Braket{\ba_i,\bb_j}}\right]_{i\in[1,n], j\in[1,n]},\\
            &
            \bA_2\coloneqq \left[\exp{B^2}\right]_{i\in[1,n],j\in[n+1,2n]},\\
            &
            \bA_3\coloneqq \left[0\right]_{i\in[n+1,2n],j\in[1,n]},\\
            &
            \bA_4\coloneqq 
            \left[\exp{B^2}\right]_{ i\in[n+1,2n],j\in[n+1,2n] }.
        \end{align*}
        We provide the explicit form of \eqref{eqn:A_matrix_compact} in \eqref{eqn:A_matrix_full}.

        For each $(i,j) \in [n]\times[n]$, it holds
        \begin{align}
            \bA_{i,j} 
            &= \exp{\beta B^2 \Braket{ \ba_i,\bb_j} } \nonumber \\
            &\leq \exp{\beta B^2 \Tilde{d} \norm{\ba_i}_{\text{max}} \norm{\bb_j}_{\text{max}}} \leq \exp{B^2}. \nonumber
        \end{align}
        Thus, 
        \begin{equation}
        0 \leq \bA_{i,j} \leq \exp{B^2}. \nonumber 
        \end{equation}
        Since $\bD = \diag\left(\bA \textbf{1}_{\Tilde{n}}\right)$, for each $i \in [\Tilde{n}]$, we get
        \begin{equation}
        \label{eqn:DBound}
            n \exp{B^2} \leq \bD_{i,i} \leq 2n \exp{B^2}.
        \end{equation} 
        
        \textbf{Scenario 2 - Part 2.}
        This part shows the $\mathtt{Gap}\text{-}\mathtt{ANNS}$ is a part of the associated $\mathtt{AHop}$ problem. 
        Given input matrices $\bD \in \R^{\Tilde{n}\times \Tilde{n}}$, $\bA \in \R^{\Tilde{n}\times \Tilde{n}}$, if we have an algorithm $\mathtt{AHop}(\Tilde{d}, \Tilde{n},\Tilde{n},\beta,B,\delta_H)$ such that its output $\Tilde{\bZ}$ satisfies
        \begin{equation}
        \label{eqn:AHop_approx}
        \norm{\Tilde{\bZ} - \bm\Xi\bD^{-1}\bA}_{\text{max}} \leq \delta_H.
        \end{equation}
        
        To connect \eqref{eqn:AHop_approx} to $\mathtt{Gap}\text{-}\mathtt{ANNS}$, we define $\Tilde{t}$ as
        \begin{align*}
            \Tilde{t} \coloneqq \frac{1}{3} \frac{\exp{\frac{1}{4} B^2(1-t/d)}}{2n\exp{B^2}}.
        \end{align*}
        It follows that 
        \begin{align*}
            \Tilde{t} &= \frac{1}{6n}\exp{-\frac{3}{4}B^2-\frac{1}{4}B^2t/d}\nonumber \\
            &= \frac{1}{6n} \exp{-\frac{3}{4} B^2 - \frac{1}{4} B^2 C_0/C} 
            \nonumber \\
            &= \frac{1}{6} \exp{-\frac{3}{4} C_\beta^2 \log n - \frac{1}{4} \frac{C_0}{C} C_\beta^2 \log n - \log n} \annot{By \eqref{def: B}} \nonumber \\
            &= \frac{1}{6} n^{-\frac{3}{4} C_\beta^2 - \frac{1}{4} \frac{C_0}{C} C_\beta^2 - 1}
            \geq n^{-C_\alpha}
            = \delta_H. 
        \end{align*}
        
        Since $\Tilde{t} \geq \delta_H$, 
        the last row vector of $\Tilde{\bZ}$, i.e  $\underline{\Tilde{\bz}}_{\Tilde{d}} \in \R^{1\times \Tilde{n}}$ for all $j\in[\Tilde{n}]$, satisfying
        \begin{equation}
        \label{eqn:ANN_Bound}
        \abs{ \underline{\Tilde{\bz}}_{\Tilde{d}}[j] - \left( 
            \underline{\bm\xi}_{\Tilde{d}} \bD^{-1}\bA \right)[j]}  
            \leq \Tilde{t},
        \end{equation}
        where $\underline{\bm\xi}_{\Tilde{d}} = \textbf{1}_{\Tilde{n}}^\sT$ is the last row of $\bm\Xi$.
        
        \textbf{Scenario 2 - Part 3.} 
        This part shows how to distinguish the 2 cases in the $\mathtt{Gap}\text{-}\mathtt{ANNS}$ with $\underline{\Tilde{\bz}}_{\Tilde{d}}$ constructed in the previous $\mathtt{AHop}(\Tilde{d}, \Tilde{n},\Tilde{n},\beta,B,\delta_H)$ problem.
        
        For the sake of convenience, we assume each input vector has an equal probability of being either $0$ or $1$, that is, 
        \begin{align*}
            \left\{
            \begin{array}{l}
                \norm{\ba_i}_2^2 = d/2, \quad \forall i \in [n], \\
                \norm{\bb_j}_2^2 = d/2, \quad \forall j \in [n].
            \end{array}
            \right.
        \end{align*}
        Hence, for each $(i,j) \in [n]\times[n]$, 
        \begin{align}
        \label{eqn:Gap}
            \beta B^2\Braket{ \ba_i, \bb_j} 
            &= \frac{B^2}{4d}(\norm{\ba_i}_2^2 + \norm{\bb_j}_2^2 - \norm{\ba_i-\bb_j}_2^2) \nonumber\\
            &= \frac{B^2}{4d} \left(d - \norm{\ba_i-\bb_j}_2^2 \right).
        \end{align}
        
        Our goal of solving $\mathtt{Gap}\text{-}\mathtt{ANNS}(d, n, t, \delta)$ is to determine, for each $i \in [n]$, whether there is a $j \in [n]$ such that $\norm{\ba_i - \bb_j}_2^2 \leq t$, or whether $\norm{\ba_i-\bb_j}_2^2 > (1+\delta)t$ for all $j\in[n]$. 
        
        \textbf{Case 1: }If there exists an $(i,j) \in [n]\times[n]$ such that $\norm{\ba_i-\bb_j}_2^2 \leq t$, then 
            \begin{align*}
                \beta B^2\Braket{\ba_i,\bb_j} \geq \frac{1}{4} B^2 (1-t/d). \annot{By \eqref{eqn:Gap}} \nonumber
            \end{align*}
            In this case,
            \begin{align*}
            \underline{\Tilde{\bz}}_{\Tilde{d}}[j]
                &\geq
                \sum^n_\iota \bD_{\iota,\iota}^{-1} \bA_{\iota,j} - \Tilde{t} \annot{By \eqref{eqn:ANN_Bound}} \nonumber \\
                &\geq \bD_{i,i}^{-1} \exp{\beta B^2\Braket{\ba_i,\bb_j}} - \Tilde{t} 
                \nonumber \\
                &\geq \frac{\exp{\frac{1}{4}B^2(1-t/d)} }{ 2n \exp{B^2} } - \Tilde{t} \annot{By \eqref{eqn:DBound}}\nonumber \\
                &= 2\Tilde{t}.
            \end{align*}
        \textbf{Case 2: }If $\norm{\ba_i-\bb_j}_2^2 > (1+\delta)t$ for all $(i,j) \in [n]\times[n]$, then 
        \begin{align}
                \beta B^2\Braket{\ba_i,\bb_j} < \frac{1}{4} B^2 (1-(1+\delta)t/d). \annot{By \eqref{eqn:Gap}} 
            \end{align}
            In this case,
            \begin{align*}
                \underline{\Tilde{\bz}}_{\Tilde{d}}[j] 
                &\leq \sum^n_\iota \bD_{\iota,\iota}^{-1} \bA_{\iota,j} +\Tilde{t} \annot{By \eqref{eqn:ANN_Bound}} \nonumber \\
                &= \sum^n_\iota \bD_{\iota,\iota}^{-1} \exp{\beta B^2\Braket{\ba_\iota,\bb_j}} + \Tilde{t} \nonumber \\
                &< \frac{ n\exp{\frac{1}{4}B^2(1-(1+\delta)t/d)}} {n \exp{B^2}} + \Tilde{t} \annot{By \eqref{eqn:DBound}}\nonumber\\
                &= \frac{\exp{\frac{1}{4}B^2(1-t/d)}}{2n\exp{B^2}} \frac{2n}{\exp{\frac{\delta}{4}B^2 t/d}} + \Tilde{t} \nonumber\\
                &= 3\Tilde{t} \cdot \frac{2n}{\exp{\frac{\delta}{4}  C_\beta^2\log n C_0/C}} + \Tilde{t}
                \nonumber \annot{By \eqref{def: C_a/C_b}} \\
                &< 2\Tilde{t}.
            \end{align*}
        Therefore, by determining whether $\underline{\Tilde{\bz}}_{\Tilde{d}}[j] \geq 2\Tilde{t}$, we distinguish the two cases, or solve the $\mathtt{Gap\text{-}Ann}(n,d,t,\delta)$. Furthermore, the entire algorithm take $T$ time, the same as the time required for $\mathtt{AHop}(\Tilde{d}, \Tilde{n},\Tilde{n},\beta,B,\delta_H)$.   
\end{proof}
\begin{corollary}
\label{thm: hardness}
    Assuming \cref{hyp:seth}, for every $q>0$, for any chosen $C, C_0>0$, there exist $\delta\in(0,0.1)$ and $C_\alpha, C_\beta>0$ satisfying \eqref{def: C_a/C_b} such that $\mathtt{AHop}(2d, M=2n,L=2n,\beta=1/2d,B=C_\beta\sqrt{\log n},\delta_H = n^{-C_\alpha})$ requires $\Omega(n^{2-q})$ time.

\end{corollary}

\begin{proof}
    By \cref{lemma:2.6}, suppose $\delta \in (0,0.1)$, $(1+\delta)\text{-}\mathtt{ANNS}$ with dimension $d=C\log n$ requires $\Omega(n^{2-q})$ time. 
    By \cref{lemma:2.7}, $\mathtt{Gap}\text{-}\mathtt{ANNS}$ requires $\calO(T + n^{2-q})$ time with $T$ being the computation time of $\mathtt{AHop}(d, M,L,\beta,B,\delta_H)$. 
    For $\mathtt{Gap}\text{-}\mathtt{ANNS}$ to have the same precision $\delta$ as $(1+\delta)\text{-}\mathtt{ANNS}$, 
    we need $\calO(T + n^{2-q}) = \Omega(n^{2-q})$.
    Consequently,
    $\mathtt{AHop}(d, M,L,\beta,B,\delta_H)$ requires $T = \Omega(n^{2-q})$ time.
    This completes the proof.
\end{proof} 
Interestingly,
\cref{thm: hardness} characterizes a phase transition behavior in $\mathtt{AHop}$ problems assuming \cref{hyp:seth}.
To extend the applicability of this corollary beyond the specific case where $M=L=n$, we introduce $\tau \coloneqq \max \{M,L\}$ to capture the larger dimension. That is, regardless of whether $M$ or $L$ is larger, $\tau$ ensures that the hardness result considers the worst-case scenario (i.e. extending the shorter one).
To sum up, we establish a criterion $B^\star=\Theta(\sqrt{\log \tau})$ for $\norm{\bm\Xi}_{\text{max}}$ and $\norm{\bX}_{\text{max}}$ such that, only below which, solving $\mathtt{AHop}$ in $\tau^{2-\Omega(1)}$ (sub-quadratic) time is possible.

\section{An Almost Linear Modern Hopfield Model}
\label{sec:method}
To showcase our theory, this section presents an example of an almost linear-time modern Hopfield model using low-degree polynomial approximation. 
We show its almost linear lower bound on computational time in \cref{sec:low_rank_AHop} and its upper bound on memory retrieval error in the same section. Additionally, we show that this model possesses a marginally smaller, yet still exponential-in-$d$ memory capacity in \cref{sec:error_bound},  compared to standard modern Hopfield associative memory models \cite{wu2023stanhop,hu2023SparseHopfield,ramsauer2020hopfield}.

\subsection{Background: Polynomial Method for Low-Rank Approximation}

Consider a matrix $\bA\in\R^{p\times q}$, and a function $f:\R\to \R$. We define $f(\bA):\R^{p\times q}\to \R^{p\times q}$ as the matrix obtained by applying $f$ entry-wise to $\bA$. The polynomial method aims to find a low-rank approximation for $f(\bA)$. Under this method, if $\bA$ possesses a low rank, and if function $f$ can be well-approximated by a low-degree polynomial, then the matrix $f(\bA)$ can be approximated by a low-rank matrix. Furthermore, this low-rank approximation can be efficiently computed in terms of its low-rank decomposition.

\citet{aggarwal2022optimal} provide the bounds on the degrees of the polynomial required for low-rank approximation of $f(\bA)$, particularly when $f$ is the exponential function. 
Leveraging these results, we construct a low-rank approximation for $\Softmax(\beta \bm\Xi^\sT \bX)$ in \eqref{eqn:MHM}, satisfying the following definition: 

\begin{definition}[$(\delta_A,r)$ Low-Rank Approximation]
\label{def:2.5}
    Let $r \in \mathbb{N}_+ \ge 1$ and $\delta_A \in(0,0.1)$.
    For a given $\bA \in \R^{p \times q}$,
    we say $\Tilde{\bA} \in \R^{p \times q}$ is an $(\delta_A,r)$-approximation of $\bA$ if
    \begin{itemize}
        \item 
        $\Tilde{\bA} = \bU \bV^\sT$ with $\bU\in\R^{p\times r}$  and $\bV\in \R^{q\times r}$, and 

        \item
        $\abs{ \Tilde{\bA}_{ij} - \bA_{ij}} \leq \delta_A\cdot \bA_{ij}$ for each $i\in [p]$ and $j\in [q]$.
    \end{itemize}
\end{definition}

\subsection{Low-Rank Matrix Approximation for $\mathtt{AHop}$ }
\label{sec:low_rank_AHop}
This section includes our linear time result for $\mathtt{AHop}$ via low-rank approximation.
Let 
$\norm{\bX}_{\text{max}}\le B$ and $\norm{\bm{\Xi}}_{\text{max}}\le B$. Let $\mathtt{T}_{\text{mat}} (a,b,c)$ denote the time required for multiplication between an $\R^{a \times b}$ matrix and an $\R^{b\times c}$ matrix. 
In fact, $\mathtt{T}_{\text{mat}} (a,b,c) \leq \calO(abc)$.

\begin{algorithm*}[!htp]
       \caption{The algorithm to solve $\mathtt{AHop}$}
       \label{alg:algorithm1}
    \begin{algorithmic}[1]
       \item[] {\bfseries Input:} matrices $\bm{\Xi}\in \R^{d\times M},\bX\in \R^{d\times L}$, with $\beta, d, M, L, B$, and error margin $\delta_A$. Let $\tau \coloneqq \max\left\{M,L\right\}$.

       \STATE $g \gets \calO\(\max\left\{B^2 \beta d,\frac{\log(1/\delta_A)}{\log\[ 1/(B^2\beta d)\cdot \log(1/\delta_A)\]} \right\}\)$ by \cref{lemma:2.4}
       
       \STATE $r \gets {2(g+d) \choose 2g}$ by \cref{lemma:2.4}
       
       \STATE Compute $\bU_1 \in \R^{M \times r}, \bU_2 \in \R^{L \times r}$ by \cref{lemma:2.3} \hfill // Time: $\calO(\tau  rg)$
       
       \STATE Compute $\Tilde{\bD}^{-1} = \diag\(\bU_1(\bU_2^{T}\mathbf{1}_L) \) \in \R^{M\times M}$ \hfill // Time: $\calO(\tau r)$
       
       \STATE $\Tilde{\bZ} \gets \bm\Xi\Tilde{\bD}^{-1} \bU_1 \bU_2^\sT \in \R^{d\times M}$ \hfill // Time: $\calO \left(\tau rd \right)$
       
       \item[] \textbf{return} $\Tilde{\bZ}$
    \end{algorithmic}
\end{algorithm*}

We compute $\Tilde{\bA}$ as a $(\delta_A,r)$-approximation of $\bA$:
\begin{lemma}
    \label{lemma:2.4}
    Suppose $B > 1$ and matrices $\bm\Xi \in \R^{d\times M}$, $\bX \in \R^{d\times L}$ have $\norm{\bX}_{\text{max}} \leq B$ and $\norm{\bm\Xi}_{\text{max}} \leq B$. 
    Given $\bA = \exp{\beta\bm\Xi^\sT\bX} \in \R^{M \times L}$, for $\delta_A\in(0,0.1)$, there is a positive integer $g$ upper bounded by
    \begin{align*}
    g = \calO\(\max\left\{B^2 \beta d,\frac{\log(1/\delta_A)}{\log\[ 1/(B^2\beta d)\cdot \log(1/\delta_A)\]} \right\}\),
    \end{align*} 
    and a $r\in \mathbb{N}_+$ upper bounded by 
    $r \leq {2(g+d) \choose 2g} $
    such that: There is a matrix $\Tilde{\bA} \in \R^{M \times L}$ that is an $(\delta_A,r)$-approximation of $\bA \in \R^{M \times L}$. Furthermore, the matrices $\bU_1$ and $\bU_2$ defining $\Tilde{\bA}$ can be computed in $\calO( \max\left\{M,L\right\} \cdot rg)$ time.
\end{lemma}
\begin{proof}
    For each $(i,j) \in [M] \times [L]$, we have
    \begin{align*}
        \left|(\bm\Xi^\sT\bX)_{i,j}\right| = \left|\sum_{l=1}^d \bm\Xi_{l,i}\bX_{l,j}\right| \leq  \norm{\bm\Xi}_{\text{max}}\norm{\bX}_{\text{max}}d \leq B^2d.
    \end{align*}
    Thus, the entries of the $\exp$ in $\bA$ have upper bound:
    \begin{align*}
        \norm{\beta\bm\Xi^\sT\bX}_{\text{max}} \leq B^2 \beta d.
    \end{align*}
    Applying \cref{lemma:2.2} with bound $B^2 \beta d$, there is a polynomial function $P(x)$ of degree $g$ such that:
    \begin{align*}
        \sup_{\bm\Xi,\bX} \abs{P((\beta\bm\Xi^\sT\bX)_{ij})-\bA_{ij}}< \delta_A \cdot \bA_{ij}.
    \end{align*}
    Applying \cref{lemma:2.3} with $\bm\Xi$, $\bX$, there exists an algorithm constructing $\bU_1, \bU_2$ in $\calO( \max\left\{M,L\right\} \cdot rg)$ time such that $P(\beta\bm\Xi^\sT\bX) = \bU_1\bU_2^\sT$.
    
    Therefore, by \cref{def:2.5}, $\Tilde{\bA} \coloneqq P(\beta\bm\Xi^\sT\bX)$ is an $(\delta_A,r)$-approximation of $\bA$.
\end{proof}
Prior to solving $\mathtt{AHop}$, we compute the approximation error bound for $\Tilde{\bZ}$ by utilizing a low-rank approximation (\cref{lemma:2.4}) applied to \cref{prob:AHop}.
\begin{lemma}[Approximation Error]
\label{lemma:approx_error}
    Let $\delta_A \in \(0,0.1\)$, $\beta>0$, $B > 0$, $\norm{\bX}_{\text{max}} \leq B$, and $\norm{\bm\Xi}_{\text{max}} \leq B$. Let $\bA = \exp{\beta\bm\Xi^\sT\bX} \in \R^{M \times L}$, and let $\Tilde{\bA} \in \R^{M \times L}$ such that, for each $(l,j) \in [M] \times [L]$,
    \begin{equation}
    \label{eqn:A_error}
        \abs{\Tilde{\bA}_{l,j} - \bA_{l,j}} \leq \delta_A \cdot \bA_{l,j}.
    \end{equation}
    
    Let $\bD=\diag(\bA\textbf{1}_M)\in \R^{M\times M}$ and $\Tilde{\bD}=\diag(\Tilde{\bA}\textbf{1}_M)\in \R^{M\times M}$, for each $l\in[M]$, we have
    \begin{equation}
    \label{eqn:D_error}
        \abs{\Tilde{\bD}_{l,l} - \bD_{l,l}} \leq \delta_A \cdot \bD_{l,l}.
    \end{equation}
    Hence, we have 
    \begin{equation}
    \label{eqn:Z_error}
        \norm{\Tilde{\bZ} - \bZ}_{\text{max}} = \norm{ \bm\Xi \Tilde{\bD}^{-1} \Tilde{\bA} - \bm\Xi\bD^{-1} \bA }_{\text{max}} \leq 2MB\delta_A.
    \end{equation}
\end{lemma}
\begin{proof}\vspace{-1em}
    To see \eqref{eqn:D_error}, we observe
    \begin{align*}
        \abs{\Tilde{\bD}_{l,l} - \bD_{l,l}} 
        &= \abs{\sum^n_{j=1}\Tilde{\bA}_{l,j} - \sum^n_{j=1}\bA_{l,j}}\\
        &\leq \sum^n_{j=1} \abs{\Tilde{\bA}_{l,j} - \bA_{l,j}} \nonumber\\
        &\leq \sum^n_{j=1} \delta_A \cdot \bA_{l,j} = \delta_A \cdot \bD_{l,l}. \annot{By \eqref{eqn:A_error}}
    \end{align*}
    By triangle inequality, we have
    \begin{align*}
    &~\norm{\bm\Xi\Tilde{\bD}^{-1}\Tilde{\bA} - \bm\Xi\bD^{-1}\bA}_{\text{max}}\\
    \leq &~ \underbrace{\norm{\bm\Xi\Tilde{\bD}^{-1}\Tilde{\bA} - \bm\Xi\bD^{-1}\Tilde{\bA}}_{\text{max}}}_{\text{(I)}} + \underbrace{\norm{\bm\Xi\bD^{-1}\Tilde{\bA} - \bm\Xi\bD^{-1}\bA}_{\text{max}}
    }_{\text{(II)}}.
    \nonumber
    \end{align*} 
    Consider the (I) term; for each $(i,j) \in [d]\times[L]$, we have
    \begin{align*}
        &~ \abs{\(\bm\Xi \Tilde{\bD}^{-1}\Tilde{\bA} - \bm\Xi \bD^{-1}\Tilde{\bA}\)_{i,j}} \\
        = &~ \abs{\sum_{l=1}^M \bm\Xi_{i,l} \(\Tilde{\bD}_{l,l}^{-1} - \bD_{l,l}^{-1}\)\Tilde{\bA}_{l,j}} 
        \nonumber\\
        \leq &~ \sum_{l=1}^M\abs{\(\Tilde{\bD}_{l,l}^{-1} - \bD_{l,l}^{-1}\)\Tilde{\bA}_{l,j}} \cdot\norm{\bm\Xi}_{\text{max}} \nonumber\\
        = &~ \sum_{l=1}^M\abs{\frac{\bD_{l,l}-\Tilde{\bD}_{l,l}}{\bD_{l,l}\Tilde{\bD}_{l,l}}\Tilde{\bA}_{l,j}} \cdot\norm{\bm\Xi}_{\text{max}} \nonumber\\
        \leq &~ \delta_A B \sum^M_{l=1} \Tilde{\bD}^{-1}_{l,l}\Tilde{\bA}_{l,j} 
        \annot{By $\norm{\bm\Xi}_{\text{max}} \leq B$ and \eqref{eqn:D_error}} \nonumber \\
        \leq &~ \delta_A B \cdot M.
    \end{align*}
    
    Consider the (II) term; for each $(i,j) \in [d]\times[L]$, we have
    \begin{align*}
        &~\abs{\(\bm\Xi\Tilde{\bD}^{-1}\Tilde{\bA} - \bm\Xi\bD^{-1}\bA\)_{i,j}} \\
        = &~ \abs{\sum_{l=1}^M \bm\Xi_{i,l} \bD_{l,l}^{-1}\(\Tilde{\bA}_{l,j} - \bA_{l,j}\)} \nonumber \\
        \leq &~ \sum_{l=1}^M\abs{\bD_{l,l}^{-1}} \cdot \abs{\(\Tilde{\bA}_{l,j} - \bA_{l,j}\)} \cdot \norm{\bm\Xi}_{\text{max}} \nonumber\\
        \leq &~ \delta_A B\sum_{l=1}^M\bD_{l,l}^{-1} \bA_{l,j} \annot{By $\norm{\bm\Xi}_{\text{max}} \leq B$ and \eqref{eqn:A_error}}\nonumber\\
        \leq &~ \delta_A B \cdot M.
    \end{align*}
    Combining (I) and (II), we obtain
    \begin{align*}
        \norm{\bm\Xi\Tilde{\bD}^{-1} \Tilde{\bA}- \bm\Xi \bD^{-1} \bA }_{\text{max}} \leq 2MB\delta_A.
    \end{align*}
    This completes the proof of $\eqref{eqn:Z_error}$.
\end{proof}
\cref{lemma:approx_error} states that the controllable approximation error in \cref{prob:AHop} takes the form of $\delta_H=2MB\delta_A$ by low-rank approximation.
Here $M$ is the size of stored memory set $\bm{\Xi}$, $\delta_A$ is the precision of low-rank approximation and $B$ is the upper bound of $\norm{\bX}_{\max}$ and $\norm{\bm{\Xi}}_{\max}$.

Next, we show that $\mathtt{AHop}$ utilizing $(\delta_A,r)$-approximation requires only almost linear computational time.

\begin{theorem}[Almost Linear $\mathtt{AHop}$, \cref{alg:algorithm1}]
\label{thm: eff_algo}
    Let $\tau \coloneqq \max\left\{M,L\right\}$ and $\delta_H \coloneqq 2MB\delta_A$. For $\beta>0$, $d,M,L\in \mathbb{N}_+$, $\delta_A >0$,  $\norm{\bX}_{\text{max}} \leq B$ and $\norm{\bm\Xi}_{\text{max}} \leq B$ with  $B\geq 1$, there are $g = \calO\(\max\left\{B^2 \beta d,\frac{\log(1/\delta_A)}{\log\[ 1/(B^2\beta d)\cdot \log(1/\delta_A)\]} \right\}\) \in \mathbb{N}_+$ and $ r ={2(g+d) \choose 2g}  \in \mathbb{N}_+$ such that: There exists an \cref{alg:algorithm1} that runs in $ \calO \left(\tau rg + \tau rd\right) $ time to solve $\mathtt{AHop}(d,M,L,\beta,B,\delta_H)$.
    Thus, under realistic settings where $d = \calO(\log \tau), \beta = \Theta(1/d), \delta_H = MB/poly(\tau)$, if $B = o(\sqrt{\log \tau})$, \cref{alg:algorithm1} requires time $\tau^{1+o(1)}$. 
\end{theorem}
\begin{proof}
    In \cref{alg:algorithm1}, step~3 requires $\calO\left(\tau rg\right)$ time by \cref{lemma:2.4}; step~4 requires $\mathtt{T}_{\text{mat}}(r,L,1) + \mathtt{T}_{\text{mat}}(M,r,1) = \calO\left(\tau r\right)$ time; step~5 requires $dM + \mathtt{T}_{\text{mat}}(d,M,r) + \mathtt{T}_{\text{mat}}(d,r,L)=\calO\left(\tau rd\right)$ time. Thus, \cref{alg:algorithm1} requires $\calO \left(\tau rg + \tau rd\right)$ time.
    
    If the parameters satisfy $d = \calO(\log \tau), \beta = \Theta(1/d),B = o(\sqrt{\log \tau})$, and $\delta_A = 1/poly(\tau) = \tau^{-\calO(1)}$, we have
    \begin{align*}
        g 
        &= \calO\(\max\left\{B^2 \beta d, \frac{\log(1/\delta_A)}{\log\[ 1/(B^2\beta d)\cdot \log(1/\delta_A)\]} \right\}\) \\
        &= \calO\(\max\left\{o(\log \tau),\frac{\log \tau}{\log(\log \tau)} \right\}\) 
        = o(\log \tau).
    \end{align*}
    We write $g$ as $\log \tau/f$ with any $f = \omega(1)$, then
    \begin{align*}
        r & = {2(d+g) \choose 2g} \leq \(\frac{e(d+g)}{g}\)^{2g} = 2^{\calO(g \log ((d+g)/g))} 
        \\
        &
        \leq 2^{\calO(g \log (\log \tau/g))} = 2^{\calO(\log \tau \log f/f)} 
        \\
        & < 2^{o(\log \tau)} 
        < \tau^{o(1)}.
    \end{align*}
    We know $(\log \tau)^{\calO(1)} \le \tau^{c}$ for all $a,c>0$ and $b>1$, so 
    \begin{align*}
        \calO(\tau^a (\log\tau)^b)\le \tau^a\cdot \tau^{o(1)}=\tau^{a+o(1)},
    \end{align*}where $\tau^{a+o(1)}$ means $\tau^{a+o(1)}$ grows slightly larger than $\tau^{a}$.
    Since $d,r,g=\calO(\log \tau)$,  there exists some constant $K$ such that $d,r,g \le K \log \tau$. Thus, \cref{alg:algorithm1} requires time:
    \begin{align*}
        \calO\left(\tau r d + \tau r g\right)
        &\leq \calO \left(\tau (\log\tau)^2\right)
        \leq \tau^{1+o(1)}.
    \end{align*}
    This completes the proof.
\end{proof}
\cref{thm: eff_algo} provides a formal example of efficient computation \cref{alg:algorithm1} for $\mathtt{AHop}$ using low-rank approximation (\cref{lemma:2.4}) within a controllable approximation error (\cref{lemma:approx_error}). 
This corresponds to \cref{thm: hardness} when the efficient criterion holds. 
Specifically, to achieve efficient computation under realistic settings, we require $B = o(\sqrt{\log \tau})$, leading to almost linear running time $\tau^{1+o(1)}$.

\subsection{Memory Retrieval Error Bound}
\label{sec:error_bound}
Considering the standard modern Hopfield retrieval dynamics with length-$L$ query sequences from \eqref{eqn:MHM}:
\begin{align*}
\bZ= \bm{\Xi}\Softmax\(\beta \bm{\Xi}^\sT \bX\).
\end{align*}
Let $\Tilde{\bZ}\in\R^{d\times L}$ be the output of the \textit{efficient} memory retrieval dynamics by \cref{alg:algorithm1} retrieving $\bX^{\text{new}}$ from stored memory set $\bm{\Xi}\in\R^{d\times M}$ based on given query $\bX\in\R^{d\times L}$.

To see how this approximate model stores and retrieves memory patterns, we first introduce the following definitions.
\begin{definition}
    Given a function $\calT:\R^d\to \R^d$.
    A generalized fixed point of $\calT$ is a point $\bx\in\R^d$ for which $\bx\in \calT(\bx)$.
\end{definition}
\begin{definition} [Memory Storage and Retrieval]
\label{def:stored_and_retrieved}
For each $\mu\in[M]$, let $R\coloneqq \half \Min_{\mu,\nu\in[M];\mu\neq\nu}\norm{\bxi_\mu-\bxi_\nu}$ be the finite radius 
 of each sphere $\calS_\mu$ centered at memory pattern $\bxi_\mu$.
We say $\bxi_\mu$ is \textit{stored} if all $\bx\in\calS_\mu$ are generalized fixed points of $\calT$, $\bx^\star_\mu \in \calS_\mu$, and $\calS_\mu \cap \calS_\nu=\emptyset$ for $\mu \neq \nu$.
We say $\bxi_\mu$ is $\epsilon$-\textit{retrieved} by $\calT$ with $\bx$ for an error $\epsilon$, if $\norm{\calT(\bx)-\bxi_\mu}\le \epsilon$.
\end{definition}

\begin{remark}
\label{remark:ok_store_ret}
A direct implication from \cref{def:stored_and_retrieved} is that the approximation error of $\mathtt{AHop}$ (see \cref{eqn:AHop_approx}) must satisfy $\delta_H = 2MB\delta_A < R$ for successful memory retrieval (and storage).
\end{remark}
Additionally, we recall the following definition regarding the separation between memory patterns.
\begin{definition}[Separation of Patterns]
\label{def:separation_of_patterns}
The separation of a memory pattern $\bxi_\mu$ from all other memory patterns $\bm{\Xi}$ is defined as its minimal inner product difference to any other patterns:
$\Delta_\mu\coloneqq
\Min_{\nu,\nu\neq \mu}\[\Braket{\bxi_\mu,\bxi_\mu}-\Braket{\bxi_\mu,\bxi_\nu}\]$.
\end{definition}
Next, we present the retrieval error bound of $\Tilde{\bZ}$.
\begin{theorem}[Retrieval Error]
\label{thm:eps_sparse_dense}
Let $\bar{\bm{\Xi}}$ be the ground truth memory sequence corresponding to $\bX$.
Suppose $\bx_{l}\in S_\mu$ with some $\mu\in[M]$ for each $l\in[L]$, it holds
\begin{align}
\label{eqn:AMHM_error}
    &\norm{\Tilde{\bZ}-\bar{\bm{\Xi}}}_{\text{max}}
    \\
    &\le
    2B(M-1) e^{-\beta \(\Braket{\bxi_\mu,\bx}-\Max_{\nu\in[M]}\Braket{\bxi_\mu,\bxi_\nu}\)}
    + 2MB\delta_A.
    \nonumber
\end{align}
\end{theorem}
\begin{proof}
We first decompose the RHS of \eqref{eqn:AMHM_error} as
    \bea
    \norm{\Tilde{\bZ}-\bar{\bm{\Xi}}}_{\text{max}}
    =
    \Big\|\underbrace{\(\Tilde{\bZ}-\bZ\)}_{\text{Approximation Error}}+\underbrace{\(\bZ-\bar{\bm{\Xi}}\)}_{\text{Retrieval Error }}\Big\|_{\text{max}}.
    \eea
Then, we bound the approximation error with \cref{lemma:approx_error} and bound the retrieval error with \citep[eqn.~2.7]{hu2023SparseHopfield}.
By triangle inequality, we complete the proof. 
\end{proof}
\begin{remark}
    By definition of $\norm{\cdot}_{\text{max}}$, this bound also holds for retrieval based on single pattern $\bx$.
\end{remark}
\begin{remark}
Similar to standard results of modern Hopfield models \cite{wu2024uniform,wu2023stanhop,hu2023SparseHopfield,ramsauer2020hopfield}, \eqref{eqn:AMHM_error} indicates that with sufficiently large $\Delta_\mu$ and sufficiently small approximation error, \cref{alg:algorithm1} retrieves memory patterns in a single \textit{iteration}. 
This allows this efficient modern Hopfield model to serve as a network layer with a single activation, enabling its integration into deep learning, similar to \cite{hu2024outlier,xu2024bishop,schimunek2023contextenriched,hoover2023energy,seidl2022improving,furst2022cloob,paischer2022history}.
\end{remark}
Surprisingly, this model achieves almost linear time efficiency while maintaining the exponential memory capacity characteristic of modern Hopfield models.
\begin{corollary}[Capacity Lower Bound, Informal]
\label{coro:capacity_informal}
Suppose all memory patterns are sampled from a sphere of radius $m$. 
This efficient modern Hopfield (approximate \eqref{eqn:MHM} with \cref{alg:algorithm1}) exhibits a  \textit{exponential-in-$d$} lower bound $M$ on the number of patterns it can store and retrieve.
\end{corollary}
\begin{proof}[Proof Sketch]
We first derive the necessary condition for a pattern to be stored and retrieved in the model, i.e., the well-separation condition.
Next, we combine it with the
separation analysis of random patterns  \cite{hu2023SparseHopfield}.
See \cref{sec:proof_capacity} for a formal version and a detailed proof.
\end{proof}
\begin{remark}
While the capacity $M$ is slightly smaller than those of \cite{wu2023stanhop,hu2023SparseHopfield,ramsauer2020hopfield}, it still scales exponentially in pattern dimension $d$.
Namely,
$\mathtt{AHop}$ as per \cref{alg:algorithm1} achieves almost linear computation time with only a marginal sacrifice in memory capacity.
\end{remark}

\section{Discussion and Conclusion}
\label{sec:conclusion}We apply the fine-grained reduction under the SETH hypothesis to study the computational limits of the retrieval dynamics of modern Hopfield associative memory models \cite{hu2024outlier,hu2024nonparametric,wu2024uniform,wu2023stanhop,hu2023SparseHopfield,ramsauer2020hopfield}.
This work holds practical significance because of the robust link between transformer attention mechanisms and modern Hopfield models.
We make a key observation by framing associative memory retrieval as an Approximate Nearest Neighbor Search (ANNS) problem, enabling the application of fine-grained reduction.
This allows us to identify a phase transition behavior on the efficiency of all possible variants of modern Hopfield models (\cref{thm: hardness}) by tuning the norm bound of queries $\bX$ and memories $\bm{\Xi}$. 
In addition, we showcase our theory with an almost linear time variant of modern Hopfield models (\cref{thm: eff_algo}).
We show this efficient model inherits the defining characteristic of modern Hopfield models: exponential memory capacity (\cref{coro:capacity_informal} and \cref{thm:memory_capacity_formal}).

\paragraph{Limitation.}
By the formal nature of this work, our results do not lead to practical implementations.
However, 
we anticipate that our findings will offer valuable insights for future efficient Hopfield-centric and transformer-based foundation models and deep learning implementations.

\clearpage
\newpage
\normalsize
\titlespacing*{\section}{0pt}{*1}{*1}
\titlespacing*{\subsection}{0pt}{*1.25}{*1.25}
\titlespacing*{\subsubsection}{0pt}{*1.5}{*1.5}

\setlength{\abovedisplayskip}{10pt}
\setlength{\abovedisplayshortskip}{10pt}
\setlength{\belowdisplayskip}{10pt}
\setlength{\belowdisplayshortskip}{10pt}
\onecolumn
\appendix
\part*{Supplementary Material}
\label{sec:append}

{
\setlength{\parskip}{-0em}
\startcontents[sections]
\printcontents[sections]{ }{1}{}
}

{
\setlength{\parskip}{-0em}
\startcontents[sections]
\printcontents[sections]{ }{1}{}
}

\section{Supplementary Theoretical Backgrounds}

\subsection{Low-Degree Approximation of $\exp$ Function}
Here we present some useful known results for later convenience.

\begin{lemma}[Approximation Degree of $e^{ x }$, Theorem~1.3 of \cite{aggarwal2022optimal}]
    For any real number $B\ge 1$ and $\delta\in(0,1)$, and function $f :[0,B]\to \R$,
    there is a polynomial function  $P:\R\to\R$ of degree tightly bounded by
    \begin{align*} 
    d_{B,\delta}(f=e^{x})
    =\Theta\(\max\left\{B,\frac{\log(1/\delta)}{\log\[1/B\cdot \log(1/\delta)\]} \right\}\),
    \end{align*}
    such that
    $\sup_{x\in[0,B]}\abs{P(x)-\exp{x}}< \delta$.
\end{lemma}
The polynomial $P(x)$ with degree $d_{B,\delta}(e^{x})$ can be computed in $\text{poly}(d_{B,\delta}(e^{x}))$ time.

\begin{lemma}[Corollary~2.2 of \cite{alman2023fast}]
    \label{lemma:2.2}
    For any real number $B\ge 1$ and $\delta\in(0,1)$, and function $f:[-B,B]\to \R$,
    there is a polynomial function  $P:\R\to\R$ of degree tightly bounded by
    \begin{align*}
    d_{B,\delta}(f=e^{x})
    =\Theta\(\max\left\{B,\frac{\log(1/\delta)}{\log\[1/B\cdot \log(1/\delta)\]} \right\}\),
    \end{align*} 
    such that
    $\sup_{x\in[0,B]}\abs{P(x)-\exp{x}}< \delta\cdot \exp{x}$.
\end{lemma}

For more related topics and techniques, please see \cite{gswy23,gsx23_incontext,syyz23_dp,rsz22} for fast approximation algorithms of attention and tensor regression via tensor trick, \cite{gu2024low} for low-rank matrix completion, \cite{pmlr-v238-song24a,dms23_rand,brand2023algorithm,swyz21} for attention kernel regression, and \cite{gll+24b,gll+24c,gu2024tensor,alman2024fine,szz24} for low-rank gradient computation in machine learning and large foundation models.

\subsection{Additional Theoretical Results: Matrix Multiplication Polynomial Approximation}

Here, we introduce a helper lemma for approximating an exponential function where the exponent involves matrix multiplication in the context of cross-attention. 
This lemma is instrumental in proving \cref{lemma:2.4}.

\begin{lemma}[Generalized from Lemma~3.2 of \cite{alman2023fast}]
\label{lemma:2.3}
Consider a polynomial function $P(x)$ representing a degree-$g$ polynomial. Given matrices $\bX \in \R^{M \times d}$ and $\bY \in \R^{L \times d}$, there exists an algorithm with a running complexity $\calO( \max\left\{M,L\right\} \cdot rg)$, where $r={2(g+d) \choose 2g}$. This algorithm, upon receiving matrices $\bX, \bY$ as input, constructs matrices $\bU_1,\bU_2$ that satisfy the equality $P(\bX \bY^\sT) = \bU_1\bU_2^\sT$, where $\bU_1 \in \R^{M \times r}$ and $\bU_2 \in \R^{L \times r}$.
\end{lemma}
\begin{proof}
    See \cref{sec:proof_poly_approx} for a detailed proof.
\end{proof}

\clearpage
\section{Proofs of Main Text}

\subsection{Proof of \cref{lemma:2.3}}
\label{sec:proof_poly_approx}
\begin{proof}
    For vectors $\bu, \bv \in \R^d$, define the union of the components $\calV \coloneqq \{u_1,\cdots,u_d,v_1,\cdots,v_d\}$. Define $\mathcal{F}$ as the set of functions $f$ such that:
    \begin{align*}
    \mathcal{F} \coloneqq \left\{ f: \calV \to \{0,1,2,\cdots,2g\} \Big| \sum_{v\in \calV} f(v) \leq 2g \right\}.
    \end{align*}
    The cardinality of $\mathcal{F}$ is derived by solving combination-with-repetition problems, leading to the expression:
    \begin{align*}
        |\mathcal{F}| = {2d+2g \choose 2g}.
    \end{align*}
    $P(x)$ can be written as:
    \begin{align*}
        P(x) = \sum^{g}_{i=0} c_i\cdot x^i.
    \end{align*}
    Let $\bu \coloneqq \left[ u_1, \cdots, u_d \right]\in\R^d$  and $\bv \coloneqq \left[ v_1, \cdots, v_d \right]\in\R^d$.     Consider the polynomial $P(\Braket{\bu,\bv})$:
    \begin{align*}
        P(\Braket{\bu,\bv}) = \sum^{g}_{i=0} c_i\cdot (\Braket{\bu,\bv})^i.
    \end{align*}
    There exists a set of constant $c_f$ associated with each function $f \in \mathcal{F}$, such that: 
    \begin{align*}
        \sum^{g}_{i=0} c_i\cdot (\Braket{\bu,\bv})^i = \sum_{f\in\mathcal{F}}c_f\cdot\prod_{v\in \calV}v^{f(v)}.
    \end{align*}
    Define two vector-valued functions $\bm{\phi}_u, \bm{\phi}_v: \mathbb{R}^d \to \mathbb{R}^{|\mathcal{F}|}$. For each $f \in \calF$, we define the elements of $\bm{\phi}_u, \bm{\phi}_v$ as follows:
    \begin{align*}
        \bm\phi_{u,f}(\bu) = c_f \cdot \prod_{l=1}^d u_l^{f(u_l)}, 
        \quad
        \bm\phi_{v,f}(\bv) = \prod_{l=1 }^d v_l^{f(v_l)}.
    \end{align*}
    Thus, $P(\Braket{\bu,\bv})$ becomes:
    \begin{align*}
        P(\Braket{\bu,\bv}) = \Braket{ \bm\phi_u(\bu),\bm\phi_v(\bv)}.
    \end{align*}
    Since $f \leq 2g$, both $\phi_{u,f}$ and $\phi_{v,f}$ require $\calO(g)$ time. Furthurmore, the inner product $\Braket{ \bm\phi_u(\bu),\bm\phi_v(\bv)}$ requires $\calO(rg)$ time, where $r = |\mathcal{F}|$.

    Consider the input matrices $\bX,\bY$. Let $\{\bx_i\}_{i\in[L]}$, $\{\by_i\}_{i\in[L]}$ be the i-th row vector of matrix $\bX$, $\bY$. 
    The polynomial can be generalized to:
    \begin{align*}
        P(\bX\bY^\sT) 
        &= \begin{bmatrix}
            P(\Braket{\bx_1,\by_1}) & P(\Braket{\bx_1,\by_2}) &\dots & P(\Braket{\bx_1,\by_L}) \\
            P(\Braket{\bx_2,\by_1}) & P(\Braket{\bx_2,\by_2}) &\dots & P(\Braket{\bx_2,\by_L}) \\
            \vdots & \vdots & \ddots & \vdots \\
            P(\Braket{\bx_M,\by_1}) & P(\Braket{\bx_M,\by_2}) &\dots & P(\Braket{\bx_M,\by_L}) 
        \end{bmatrix} \nonumber \\ 
        &= \begin{bmatrix}
            \Braket{\bm\phi_u(\bx_1),\bm\phi_v(\by_1)} & \Braket{\bm\phi_u(\bx_1),\bm\phi_v(\by_2)} &\dots & \Braket{\bm\phi_u(\bx_1),\bm\phi_v(\by_L)} \\
            \Braket{\bm\phi_u(\bx_2),\bm\phi_v(\by_1)} & \Braket{\bm\phi_u(\bx_2),\bm\phi_v(\by_2)} &\dots & \Braket{\bm\phi_u(\bx_2),\bm\phi_v(\by_L)} \\
            \vdots & \vdots & \ddots & \vdots \\
            \Braket{\bm\phi_u(\bx_M),\bm\phi_v(\by_1)} & \Braket{\bm\phi_u(\bx_M),\bm\phi_v(\by_2)} &\dots & \Braket{\bm\phi_u(\bx_M),\bm\phi_v(\by_L)}
        \end{bmatrix}.
    \end{align*}
    Therefore, we can constuct matrices $\bU_1  \in \R^{M\times |\mathcal{F}|}$ and $\bU_2  \in \R^{L\times |\mathcal{F}|}$ as follows:
    \begin{align*}
    \bU_1 &= \begin{bmatrix} \bm\phi_u(\bX_1) \ \bm\phi_u(\bX_2) \ \cdots \ \bm\phi_u(\bX_M) \end{bmatrix}^\sT, \\
    \bU_2 &= \begin{bmatrix} \bm\phi_v(\bY_1) \ \bm\phi_v(\bY_2) \ \cdots \ \bm\phi_v(\bY_L) \end{bmatrix}^\sT.
    \end{align*}
    It's trivial to observe $P(\bX\bY^\sT) = \bU_1\bU_2^\sT$. Moreover, constructing $\bU_1, \bU_2$ require time $\calO( \max\left\{M,L\right\} \cdot rg)$.
\end{proof}

\subsection{$\bA$ Matrix in Proof of \cref{lemma:2.7}}

    \begin{align}
    \label{eqn:A_matrix_full}
    \bA &= \begin{bmatrix}
                \exp{\frac{B^2}{\Tilde{d}} \langle a_1,b_1\rangle} & \exp{\frac{B^2}{\Tilde{d}} \langle a_1,b_2\rangle} & \cdots & \exp{\frac{B^2}{\Tilde{d}} \langle a_1,b_n\rangle} & \exp{B^2} & \exp{B^2} & \cdots & \exp{B^2} \\
                \exp{ \frac{B^2}{\Tilde{d}} \langle a_2,b_1\rangle } & \exp{\frac{B^2}{\Tilde{d}} \langle a_2,b_2\rangle} & \cdots & \exp{\frac{B^2}{\Tilde{d}} \langle a_2,b_n\rangle} & \exp{B^2} & \exp{B^2} & \cdots & \exp{B^2} \\
                \vdots & \vdots & \ddots & \vdots & \vdots & \vdots & \ddots & \vdots\\
                \exp{\frac{B^2}{\Tilde{d}} \langle a_n,b_1\rangle} & \exp{\frac{B^2}{\Tilde{d}} \langle a_n,b_2\rangle} & \cdots & \exp{\frac{B^2}{\Tilde{d}} \langle a_n,b_n\rangle} & \exp{B^2} & \exp{B^2} & \cdots & \exp{B^2} \\
                0 & 0 & \cdots & 0 & \exp{B^2} & \exp{B^2} & \cdots & \exp{B^2} \\
                0 & 0 & \cdots & 0 & \exp{B^2} & \exp{B^2} & \cdots & \exp{B^2} \\
                \vdots & \vdots & \ddots & \vdots & \vdots & \vdots & \ddots & \vdots\\
                0 & 0 & \cdots & 0 & \exp{B^2} & \exp{B^2} & \cdots & \exp{B^2} \\
            \end{bmatrix}.
        \end{align}

\subsection{Formal Statement and Proof of \cref{coro:capacity_informal}}
\label{sec:proof_capacity}
Let $\delta\coloneqq - 2MB\delta_A\le 0$.
\begin{theorem}[Memory Capacity Lower Bound, Formal]
\label{thm:memory_capacity_formal}
Suppose the probability of successfully storing and retrieving memory pattern is given by $1-p$.
The number of memory patterns sampled from a sphere of radius $m$ that the \\textit{efficient} modern Hopfield model (approximate \eqref{eqn:MHM} with \cref{alg:algorithm1}) can store and retrieve has a lower bound:
$M \ge \sqrt{p}C^\frac{d-1}{4}$,
where $C$ is the solution for $C=\nicefrac{b}{W_0({\exp\{a+\ln{b}\}})}$ with $W_0(\cdot)$ being the principal branch of Lambert $W$ function, %
$a\coloneqq \nicefrac{4}{d-1}\big\{\ln\[\nicefrac{2m(\sqrt{p}-1)}{(R- 2MB\delta_A)}\]+1\big\}$ and $b\coloneqq \nicefrac{4m^2\beta}{5(d-1)}$.
\end{theorem}
\begin{remark}
For details and background of Lambert $W$ function, we refer the readers to \cite{olver2010nist}.
\end{remark}

Before the main proof, we introduce the following helper lemma.
Let $m\coloneqq \Max_{\mu\in[M]} \norm{\bxi_\mu}$.
\begin{lemma}\label{thm:well_separation_condition}
Then, the {well-separation} condition of memory patterns is:
\bea
\label{eqn:separation}
\Delta_\mu \ge 
\frac{1}{\beta}\ln(\frac{2(M-1)m}{R-2MB\delta_A})+2mR.
\eea
If $2MB\delta_A=0$, \eqref{eqn:separation}  reduces to {well-separation} condition of Softmax-based  Hopfield model \cite{ramsauer2020hopfield}.
\end{lemma}

\begin{proof}[Proof of \cref{thm:well_separation_condition}]
Let $\calT_{\text{Dense}}$ be the retrieval dynamics given by the dense modern Hopfield model \cite{ramsauer2020hopfield},
and $\norm{\calT(\bx)-\bxi_\mu}$ and $\norm{\calT_{\text{Dense}}(\bx)-\bxi_\mu}$ be the approximated efficient and dense modern Hopfield model, respectively.

By \citep[Lemma~A.4]{ramsauer2020hopfield}, we have
\begin{align*}
&~\norm{\calT_{\text{Dense}}(\bx)-\bxi_\mu}
\\
\leq &~
2m(M-1)\exp{-\beta \(\Braket{\bxi_\mu,\bx}-\Max_{\nu\in[M]}\Braket{\bxi_\mu,\bxi_\nu}\)},\nonumber
\\
\leq &~
2m(M-1) \exp{-\beta\(\Delta_\mu-2mR\)},
\nonumber
\end{align*}
where $R$ is radius of the sphere $S_{\mu}$.

By \cref{thm:eps_sparse_dense}, the retrieval error $\norm{\calT(\bx)-\bxi_\mu}$ has an upper bound:
\begin{align*}
&\norm{\calT(\bx)-\bxi_\mu} 
\le 2(M-1) \exp{-\beta\(\Delta_\mu-2mR +\delta\)}m - \delta
\end{align*}
Therefore, for $\calT$ to be a mapping $\calT:S_\mu\to S_\mu$, we need 
\begin{align*}
 2(M-1) \exp{-\beta\(\Delta_\mu-2mR +\delta\)}m - \delta
\le R
\end{align*}
This deduces the well-separation condition for this almost linear time model
\begin{align*}
\Delta_\mu
\geq
\frac{1}{\beta}\ln(\frac{2(M-1)m}{R-2MB\delta_A})+2mR.
\end{align*}
This completes the proof.
\end{proof}

Now we start the main proof of \cref{thm:memory_capacity_formal}.

\begin{proof}[Proof of \cref{thm:memory_capacity_formal}]
We first observe that \eqref{thm:well_separation_condition} has a slightly tighter lower bound compared to its original counterpart \cite{ramsauer2020hopfield}, we note that under the condition identified in \cref{remark:ok_store_ret}, the new well-separation condition \cref{thm:well_separation_condition} features a smaller denominator inside the logarithmic term. 
Following a similar approach to that in \citep[Lemma~3.4]{wu2023stanhop}, we complete the proof and obtain a slightly smaller, yet still exponential-in-$d$, memory capacity lower bound.
This is an expected consequence of an efficient-accuracy tradeoff.
\end{proof}

\newpage
\twocolumn
\section*{Impact Statement}
This theoretical work, as outlined in the introduction and related works, aims to elucidate the foundations of large Hopfield- and transformer-based foundation models and is not expected to have negative social impacts.

\section*{Acknowledgments}
JH would like to thank  Feng Ruan, Dino Feng and Andrew Chen for enlightening discussions on related topics, and the Red Maple Family for support.
TL would like to thank Gamma Paradigm Research and NTU ABC-Labs for support.
The authors would like to thank the anonymous reviewers and program chairs for constructive comments.
JH is partially supported by the Walter P. Murphy Fellowship.
HL is partially supported by NIH R01LM1372201, NSF CAREER1841569, DOE DE-AC02-07CH11359, DOE LAB 20-2261 and a NSF TRIPODS1740735. 
The content is solely the responsibility of the authors and does not necessarily represent the official
views of the funding agencies.

\def\arxivfont{\rm}
\bibliographystyle{plainnat}

\bibliography{refs}

\begin{thebibliography}{64}
\providecommand{\natexlab}[1]{#1}
\providecommand{\url}[1]{\texttt{#1}}
\expandafter\ifx\csname urlstyle\endcsname\relax
  \providecommand{\doi}[1]{doi: #1}\else
  \providecommand{\doi}{doi: \begingroup \urlstyle{rm}\Url}\fi

\bibitem[Aggarwal and Alman(2022)]{aggarwal2022optimal}
Amol Aggarwal and Josh Alman.
\newblock Optimal-degree polynomial approximations for exponentials and gaussian kernel density estimation.
\newblock In \emph{Proceedings of the 37th Computational Complexity Conference}, CCC '22, Dagstuhl, DEU, 2022. Schloss Dagstuhl--Leibniz-Zentrum fuer Informatik.
\newblock ISBN 9783959772419.
\newblock \doi{10.4230/LIPIcs.CCC.2022.22}.
\newblock URL \url{https://doi.org/10.4230/LIPIcs.CCC.2022.22}.

\bibitem[Alman and Song(2023)]{alman2023fast}
Josh Alman and Zhao Song.
\newblock Fast attention requires bounded entries.
\newblock In \emph{Thirty-seventh Conference on Neural Information Processing Systems (NeurIPS)}, 2023.
\newblock URL \url{https://openreview.net/forum?id=KOVWXcrFIK}.

\bibitem[Alman and Song(2024{\natexlab{a}})]{alman2024fine}
Josh Alman and Zhao Song.
\newblock The fine-grained complexity of gradient computation for training large language models.
\newblock \emph{arXiv preprint arXiv:2402.04497}, 2024{\natexlab{a}}.
\newblock URL \url{https://arxiv.org/abs/2402.04497}.

\bibitem[Alman and Song(2024{\natexlab{b}})]{as23_tensor}
Josh Alman and Zhao Song.
\newblock How to capture higher-order correlations? generalizing matrix softmax attention to kronecker computation.
\newblock In \emph{The Twelfth International Conference on Learning Representations (ICLR)}, 2024{\natexlab{b}}.
\newblock URL \url{https://openreview.net/forum?id=v0zNCwwkaV}.

\bibitem[Alman et~al.(2020)Alman, Chu, Schild, and Song]{alman2020algorithms}
Josh Alman, Timothy Chu, Aaron Schild, and Zhao Song.
\newblock Algorithms and hardness for linear algebra on geometric graphs.
\newblock In \emph{2020 IEEE 61st Annual Symposium on Foundations of Computer Science (FOCS)}, pages 541--552. IEEE, 2020.
\newblock URL \url{https://arxiv.org/abs/2011.02466}.

\bibitem[Alman et~al.(2023)Alman, Liang, Song, Zhang, and Zhuo]{alman2023bypass}
Josh Alman, Jiehao Liang, Zhao Song, Ruizhe Zhang, and Danyang Zhuo.
\newblock Bypass exponential time preprocessing: Fast neural network training via weight-data correlation preprocessing.
\newblock In \emph{Thirty-seventh Conference on Neural Information Processing Systems (NeurIPS)}, 2023.
\newblock URL \url{https://openreview.net/forum?id=ZqSx5vXOgC}.

\bibitem[Arya et~al.(1998)Arya, Mount, Netanyahu, Silverman, and Wu]{arya1998optimal}
Sunil Arya, David~M Mount, Nathan~S Netanyahu, Ruth Silverman, and Angela~Y Wu.
\newblock An optimal algorithm for approximate nearest neighbor searching fixed dimensions.
\newblock \emph{Journal of the ACM (JACM)}, 45\penalty0 (6):\penalty0 891--923, 1998.

\bibitem[Auer et~al.(2024)Auer, Gauch, Klotz, and Hochreiter]{auer2023conformal}
Andreas Auer, Martin Gauch, Daniel Klotz, and Sepp Hochreiter.
\newblock Conformal prediction for time series with modern hopfield networks.
\newblock \emph{Advances in Neural Information Processing Systems (NeurIPS)}, 36, 2024.
\newblock URL \url{https://arxiv.org/abs/2303.12783}.

\bibitem[Backurs et~al.(2017)Backurs, Indyk, and Schmidt]{bis17}
Arturs Backurs, Piotr Indyk, and Ludwig Schmidt.
\newblock On the fine-grained complexity of empirical risk minimization: Kernel methods and neural networks.
\newblock \emph{Advances in Neural Information Processing Systems (NeurIPS)}, 30, 2017.
\newblock URL \url{https://arxiv.org/abs/1704.02958}.

\bibitem[Bommasani et~al.(2021)Bommasani, Hudson, Adeli, Altman, Arora, von Arx, Bernstein, Bohg, Bosselut, Brunskill, et~al.]{bommasani2021opportunities}
Rishi Bommasani, Drew~A Hudson, Ehsan Adeli, Russ Altman, Simran Arora, Sydney von Arx, Michael~S Bernstein, Jeannette Bohg, Antoine Bosselut, Emma Brunskill, et~al.
\newblock On the opportunities and risks of foundation models.
\newblock \emph{arXiv preprint arXiv:2108.07258}, 2021.
\newblock URL \url{https://arxiv.org/abs/2108.07258}.

\bibitem[Brand et~al.(2023)Brand, Song, and Zhou]{brand2023algorithm}
Jan van~den Brand, Zhao Song, and Tianyi Zhou.
\newblock Algorithm and hardness for dynamic attention maintenance in large language models.
\newblock \emph{arXiv preprint arXiv:2304.02207}, 2023.
\newblock URL \url{https://arxiv.org/abs/2304.02207}.

\bibitem[Brandstetter(2021)]{hopfieldblog2021}
Johannes Brandstetter.
\newblock Blog post: Hopfield networks is all you need, 2021.
\newblock URL \url{https://ml-jku.github.io/hopfield-layers/}.
\newblock Accessed: April 4, 2023.

\bibitem[Brown et~al.(2020)Brown, Mann, Ryder, Subbiah, Kaplan, Dhariwal, Neelakantan, Shyam, Sastry, Askell, et~al.]{brown2020language}
Tom Brown, Benjamin Mann, Nick Ryder, Melanie Subbiah, Jared~D Kaplan, Prafulla Dhariwal, Arvind Neelakantan, Pranav Shyam, Girish Sastry, Amanda Askell, et~al.
\newblock Language models are few-shot learners.
\newblock \emph{Advances in neural information processing systems (NeurIPS)}, 33:\penalty0 1877--1901, 2020.
\newblock URL \url{https://arxiv.org/abs/2005.14165}.

\bibitem[Burns and Fukai(2023)]{burns2023simplicial}
Thomas~F Burns and Tomoki Fukai.
\newblock Simplicial hopfield networks.
\newblock In \emph{The Eleventh International Conference on Learning Representations (ICLR)}, 2023.
\newblock URL \url{https://openreview.net/forum?id=_QLsH8gatwx}.

\bibitem[Cygan et~al.(2016)Cygan, Dell, Lokshtanov, Marx, Nederlof, Okamoto, Paturi, Saurabh, and Wahlstr{\"o}m]{cygan2016problems}
Marek Cygan, Holger Dell, Daniel Lokshtanov, D{\'a}niel Marx, Jesper Nederlof, Yoshio Okamoto, Ramamohan Paturi, Saket Saurabh, and Magnus Wahlstr{\"o}m.
\newblock On problems as hard as cnf-sat.
\newblock \emph{ACM Transactions on Algorithms (TALG)}, 12\penalty0 (3):\penalty0 1--24, 2016.
\newblock URL \url{https://arxiv.org/abs/1112.2275}.

\bibitem[Demaine(2014)]{demaine2014algorithmic}
Erik Demaine.
\newblock Algorithmic lower bounds: Fun with hardness proofs, 2014.

\bibitem[Demircigil et~al.(2017)Demircigil, Heusel, L{\"o}we, Upgang, and Vermet]{demircigil2017model}
Mete Demircigil, Judith Heusel, Matthias L{\"o}we, Sven Upgang, and Franck Vermet.
\newblock On a model of associative memory with huge storage capacity.
\newblock \emph{Journal of Statistical Physics}, 168:\penalty0 288--299, 2017.
\newblock URL \url{https://arxiv.org/abs/1702.01929}.

\bibitem[Deng et~al.(2023)Deng, Mahadevan, and Song]{dms23_rand}
Yichuan Deng, Sridhar Mahadevan, and Zhao Song.
\newblock Randomized and deterministic attention sparsification algorithms for over-parameterized feature dimension.
\newblock \emph{arXiv preprint arXiv:2304.04397}, 2023.
\newblock URL \url{https://arxiv.org/abs/2304.04397}.

\bibitem[Floridi and Chiriatti(2020)]{floridi2020gpt}
Luciano Floridi and Massimo Chiriatti.
\newblock Gpt-3: Its nature, scope, limits, and consequences.
\newblock \emph{Minds and Machines}, 30:\penalty0 681--694, 2020.

\bibitem[F{\"u}rst et~al.(2022)F{\"u}rst, Rumetshofer, Lehner, Tran, Tang, Ramsauer, Kreil, Kopp, Klambauer, Bitto, et~al.]{furst2022cloob}
Andreas F{\"u}rst, Elisabeth Rumetshofer, Johannes Lehner, Viet~T Tran, Fei Tang, Hubert Ramsauer, David Kreil, Michael Kopp, G{\"u}nter Klambauer, Angela Bitto, et~al.
\newblock Cloob: Modern hopfield networks with infoloob outperform clip.
\newblock \emph{Advances in neural information processing systems (NeurIPS)}, 35:\penalty0 20450--20468, 2022.
\newblock URL \url{https://arxiv.org/abs/2110.11316}.

\bibitem[Gao et~al.(2023{\natexlab{a}})Gao, Song, Wang, and Yin]{gswy23}
Yeqi Gao, Zhao Song, Weixin Wang, and Junze Yin.
\newblock A fast optimization view: Reformulating single layer attention in llm based on tensor and svm trick, and solving it in matrix multiplication time.
\newblock \emph{arXiv preprint arXiv:2309.07418}, 2023{\natexlab{a}}.
\newblock URL \url{https://arxiv.org/abs/2309.07418}.

\bibitem[Gao et~al.(2023{\natexlab{b}})Gao, Song, and Xie]{gsx23_incontext}
Yeqi Gao, Zhao Song, and Shenghao Xie.
\newblock In-context learning for attention scheme: from single softmax regression to multiple softmax regression via a tensor trick.
\newblock \emph{arXiv preprint arXiv:2307.02419}, 2023{\natexlab{b}}.
\newblock URL \url{https://arxiv.org/abs/2307.02419}.

\bibitem[Gu et~al.(2024{\natexlab{a}})Gu, Li, Liang, Shi, and Song]{gll+24b}
Jiuxiang Gu, Chenyang Li, Yingyu Liang, Zhenmei Shi, and Zhao Song.
\newblock Exploring the frontiers of softmax: Provable optimization, applications in diffusion model, and beyond.
\newblock \emph{arXiv preprint arXiv:2405.03251}, 2024{\natexlab{a}}.
\newblock URL \url{https://arxiv.org/abs/2405.03251}.

\bibitem[Gu et~al.(2024{\natexlab{b}})Gu, Liang, Liu, Shi, Song, and Yin]{gll+24c}
Jiuxiang Gu, Yingyu Liang, Heshan Liu, Zhenmei Shi, Zhao Song, and Junze Yin.
\newblock Conv-basis: A new paradigm for efficient attention inference and gradient computation in transformers.
\newblock \emph{arXiv preprint arXiv:2405.05219}, 2024{\natexlab{b}}.
\newblock URL \url{https://arxiv.org/abs/2405.05219}.

\bibitem[Gu et~al.(2024{\natexlab{c}})Gu, Liang, Shi, Song, and Zhou]{gu2024tensor}
Jiuxiang Gu, Yingyu Liang, Zhenmei Shi, Zhao Song, and Yufa Zhou.
\newblock Tensor attention training: Provably efficient learning of higher-order transformers.
\newblock \emph{arXiv preprint arXiv:2405.16411}, 2024{\natexlab{c}}.
\newblock URL \url{https://arxiv.org/abs/2405.16411}.

\bibitem[Gu et~al.(2024{\natexlab{d}})Gu, Song, Yin, and Zhang]{gu2024low}
Yuzhou Gu, Zhao Song, Junze Yin, and Lichen Zhang.
\newblock Low rank matrix completion via robust alternating minimization in nearly linear time.
\newblock In \emph{The Twelfth International Conference on Learning Representations (ICLR)}, 2024{\natexlab{d}}.
\newblock URL \url{https://openreview.net/forum?id=N0gT4A0jNV}.

\bibitem[Hofmann et~al.(2024)Hofmann, Schmid, Lehner, Klotz, and Hochreiter]{hofmann2024energy}
Claus Hofmann, Simon Schmid, Bernhard Lehner, Daniel Klotz, and Sepp Hochreiter.
\newblock Energy-based hopfield boosting for out-of-distribution detection.
\newblock \emph{arXiv preprint arXiv:2405.08766}, 2024.

\bibitem[Hoover et~al.(2023)Hoover, Liang, Pham, Panda, Strobelt, Chau, Zaki, and Krotov]{hoover2023energy}
Benjamin Hoover, Yuchen Liang, Bao Pham, Rameswar Panda, Hendrik Strobelt, Duen~Horng Chau, Mohammed~J Zaki, and Dmitry Krotov.
\newblock Energy transformer.
\newblock \emph{arXiv preprint arXiv:2302.07253}, 2023.
\newblock URL \url{https://arxiv.org/abs/2302.07253}.

\bibitem[Hopfield(1982)]{hopfield1982neural}
John~J Hopfield.
\newblock Neural networks and physical systems with emergent collective computational abilities.
\newblock \emph{Proceedings of the national academy of sciences}, 79\penalty0 (8):\penalty0 2554--2558, 1982.

\bibitem[Hopfield(1984)]{hopfield1984neurons}
John~J Hopfield.
\newblock Neurons with graded response have collective computational properties like those of two-state neurons.
\newblock \emph{Proceedings of the national academy of sciences}, 81\penalty0 (10):\penalty0 3088--3092, 1984.

\bibitem[Hu et~al.(2023)Hu, Yang, Wu, Xu, Chen, and Liu]{hu2023SparseHopfield}
Jerry Yao-Chieh Hu, Donglin Yang, Dennis Wu, Chenwei Xu, Bo-Yu Chen, and Han Liu.
\newblock On sparse modern hopfield model.
\newblock In \emph{Thirty-seventh Conference on Neural Information Processing Systems (NeurIPS)}, 2023.
\newblock URL \url{https://arxiv.org/abs/2309.12673}.

\bibitem[Hu et~al.(2024{\natexlab{a}})Hu, Chang, Luo, Chen, Li, Wang, and Liu]{hu2024outlier}
Jerry Yao-Chieh Hu, Pei-Hsuan Chang, Robin Luo, Hong-Yu Chen, Weijian Li, Wei-Po Wang, and Han Liu.
\newblock Outlier-efficient hopfield layers for large transformer-based models.
\newblock In \emph{Forty-first International Conference on Machine Learning (ICML)}, 2024{\natexlab{a}}.
\newblock URL \url{https://arxiv.org/abs/2404.03828}.

\bibitem[Hu et~al.(2024{\natexlab{b}})Hu, Chen, Wu, Ruan, and Liu]{hu2024nonparametric}
Jerry Yao-Chieh Hu, Bo-Yu Chen, Dennis Wu, Feng Ruan, and Han Liu.
\newblock Nonparametric modern hopfield models.
\newblock \emph{arXiv preprint arXiv:2404.03900}, 2024{\natexlab{b}}.
\newblock URL \url{https://arxiv.org/abs/2404.03900}.

\bibitem[Impagliazzo and Paturi(2001)]{ip01}
Russell Impagliazzo and Ramamohan Paturi.
\newblock On the complexity of k-sat.
\newblock \emph{Journal of Computer and System Sciences}, 62\penalty0 (2):\penalty0 367--375, 2001.

\bibitem[Indyk and Motwani(1998)]{indyk1998approximate}
Piotr Indyk and Rajeev Motwani.
\newblock Approximate nearest neighbors: towards removing the curse of dimensionality.
\newblock In \emph{Proceedings of the thirtieth annual ACM symposium on Theory of computing}, pages 604--613, 1998.

\bibitem[Ji et~al.(2021)Ji, Zhou, Liu, and Davuluri]{ji2021dnabert}
Yanrong Ji, Zhihan Zhou, Han Liu, and Ramana~V Davuluri.
\newblock Dnabert: pre-trained bidirectional encoder representations from transformers model for dna-language in genome.
\newblock \emph{Bioinformatics}, 37\penalty0 (15):\penalty0 2112--2120, 2021.

\bibitem[Kozachkov et~al.(2022)Kozachkov, Kastanenka, and Krotov]{kozachkov2022building}
Leo Kozachkov, Ksenia~V Kastanenka, and Dmitry Krotov.
\newblock Building transformers from neurons and astrocytes.
\newblock \emph{bioRxiv}, pages 2022--10, 2022.

\bibitem[Krotov and Hopfield(2016)]{krotov2016dense}
Dmitry Krotov and John~J Hopfield.
\newblock Dense associative memory for pattern recognition.
\newblock \emph{Advances in Neural Information Processing Systems (NeurIPS)}, 29, 2016.
\newblock URL \url{https://arxiv.org/abs/1606.01164}.

\bibitem[Krotov and Hopfield(2021)]{krotov2021large}
Dmitry Krotov and John~J. Hopfield.
\newblock Large associative memory problem in neurobiology and machine learning.
\newblock In \emph{International Conference on Learning Representations (ICLR)}, 2021.
\newblock URL \url{https://openreview.net/forum?id=X4y_10OX-hX}.

\bibitem[Li et~al.(2019)Li, Zhang, Sun, Wang, Li, Zhang, and Lin]{li2019approximate}
Wen Li, Ying Zhang, Yifang Sun, Wei Wang, Mingjie Li, Wenjie Zhang, and Xuemin Lin.
\newblock Approximate nearest neighbor search on high dimensional data-experiments, analyses, and improvement.
\newblock \emph{IEEE Transactions on Knowledge and Data Engineering}, 32\penalty0 (8):\penalty0 1475--1488, 2019.
\newblock URL \url{https://arxiv.org/abs/1610.02455}.

\bibitem[Moor et~al.(2023)Moor, Banerjee, Abad, Krumholz, Leskovec, Topol, and Rajpurkar]{moor2023foundation}
Michael Moor, Oishi Banerjee, Zahra Shakeri~Hossein Abad, Harlan~M Krumholz, Jure Leskovec, Eric~J Topol, and Pranav Rajpurkar.
\newblock Foundation models for generalist medical artificial intelligence.
\newblock \emph{Nature}, 616\penalty0 (7956):\penalty0 259--265, 2023.

\bibitem[Muja and Lowe(2014)]{muja2014scalable}
Marius Muja and David~G Lowe.
\newblock Scalable nearest neighbor algorithms for high dimensional data.
\newblock \emph{IEEE transactions on pattern analysis and machine intelligence}, 36\penalty0 (11):\penalty0 2227--2240, 2014.

\bibitem[Olver et~al.(2010)Olver, Lozier, Boisvert, and Clark]{olver2010nist}
Frank~WJ Olver, Daniel~W Lozier, Ronald~F Boisvert, and Charles~W Clark.
\newblock \emph{NIST handbook of mathematical functions hardback and CD-ROM}.
\newblock Cambridge university press, 2010.

\bibitem[Paischer et~al.(2022)Paischer, Adler, Patil, Bitto-Nemling, Holzleitner, Lehner, Eghbal-Zadeh, and Hochreiter]{paischer2022history}
Fabian Paischer, Thomas Adler, Vihang Patil, Angela Bitto-Nemling, Markus Holzleitner, Sebastian Lehner, Hamid Eghbal-Zadeh, and Sepp Hochreiter.
\newblock History compression via language models in reinforcement learning.
\newblock In \emph{International Conference on Machine Learning (ICML)}, pages 17156--17185. PMLR, 2022.
\newblock URL \url{https://arxiv.org/abs/2205.12258}.

\bibitem[Ramsauer et~al.(2021)Ramsauer, Sch{\"a}fl, Lehner, Seidl, Widrich, Gruber, Holzleitner, Adler, Kreil, Kopp, Klambauer, Brandstetter, and Hochreiter]{ramsauer2020hopfield}
Hubert Ramsauer, Bernhard Sch{\"a}fl, Johannes Lehner, Philipp Seidl, Michael Widrich, Lukas Gruber, Markus Holzleitner, Thomas Adler, David Kreil, Michael~K Kopp, G{\"u}nter Klambauer, Johannes Brandstetter, and Sepp Hochreiter.
\newblock Hopfield networks is all you need.
\newblock In \emph{International Conference on Learning Representations (ICLR)}, 2021.
\newblock URL \url{https://openreview.net/forum?id=tL89RnzIiCd}.

\bibitem[Reddy et~al.(2022)Reddy, Song, and Zhang]{rsz22}
Aravind Reddy, Zhao Song, and Lichen Zhang.
\newblock Dynamic tensor product regression.
\newblock In Alice~H. Oh, Alekh Agarwal, Danielle Belgrave, and Kyunghyun Cho, editors, \emph{Advances in Neural Information Processing Systems (NeurIPS)}, 2022.
\newblock URL \url{https://openreview.net/forum?id=hUjMhflYvGc}.

\bibitem[Rubinstein(2018)]{r18}
Aviad Rubinstein.
\newblock Hardness of approximate nearest neighbor search.
\newblock In \emph{Proceedings of the 50th annual ACM SIGACT symposium on theory of computing (STOC)}, pages 1260--1268, 2018.
\newblock URL \url{https://arxiv.org/abs/1803.00904}.

\bibitem[Schimunek et~al.(2023)Schimunek, Seidl, Friedrich, Kuhn, Rippmann, Hochreiter, and Klambauer]{schimunek2023contextenriched}
Johannes Schimunek, Philipp Seidl, Lukas Friedrich, Daniel Kuhn, Friedrich Rippmann, Sepp Hochreiter, and G{\"u}nter Klambauer.
\newblock Context-enriched molecule representations improve few-shot drug discovery.
\newblock In \emph{The Eleventh International Conference on Learning Representations (ICLR)}, 2023.
\newblock URL \url{https://openreview.net/forum?id=XrMWUuEevr}.

\bibitem[Seidl et~al.(2022)Seidl, Renz, Dyubankova, Neves, Verhoeven, Wegner, Segler, Hochreiter, and Klambauer]{seidl2022improving}
Philipp Seidl, Philipp Renz, Natalia Dyubankova, Paulo Neves, Jonas Verhoeven, Jorg~K Wegner, Marwin Segler, Sepp Hochreiter, and Gunter Klambauer.
\newblock Improving few-and zero-shot reaction template prediction using modern hopfield networks.
\newblock \emph{Journal of chemical information and modeling}, 62\penalty0 (9):\penalty0 2111--2120, 2022.

\bibitem[Singhal et~al.(2023)Singhal, Azizi, Tu, Mahdavi, Wei, Chung, Scales, Tanwani, Cole-Lewis, Pfohl, et~al.]{singhal2023large}
Karan Singhal, Shekoofeh Azizi, Tao Tu, S~Sara Mahdavi, Jason Wei, Hyung~Won Chung, Nathan Scales, Ajay Tanwani, Heather Cole-Lewis, Stephen Pfohl, et~al.
\newblock Large language models encode clinical knowledge.
\newblock \emph{Nature}, 620\penalty0 (7972):\penalty0 172--180, 2023.
\newblock URL \url{https://arxiv.org/abs/2212.13138}.

\bibitem[Song et~al.(2021)Song, Woodruff, Yu, and Zhang]{swyz21}
Zhao Song, David Woodruff, Zheng Yu, and Lichen Zhang.
\newblock Fast sketching of polynomial kernels of polynomial degree.
\newblock In \emph{International Conference on Machine Learning (ICML)}, pages 9812--9823. PMLR, 2021.
\newblock URL \url{https://arxiv.org/abs/2108.09420}.

\bibitem[Song et~al.(2023)Song, Yang, Yang, and Zhang]{syyz23_dp}
Zhao Song, Xin Yang, Yuanyuan Yang, and Lichen Zhang.
\newblock Sketching meets differential privacy: fast algorithm for dynamic kronecker projection maintenance.
\newblock In \emph{International Conference on Machine Learning (ICML)}, pages 32418--32462. PMLR, 2023.
\newblock URL \url{https://arxiv.org/abs/2210.11542}.

\bibitem[Song et~al.(2024{\natexlab{a}})Song, Yin, and Zhang]{pmlr-v238-song24a}
Zhao Song, Junze Yin, and Lichen Zhang.
\newblock Solving attention kernel regression problem via pre-conditioner.
\newblock In Sanjoy Dasgupta, Stephan Mandt, and Yingzhen Li, editors, \emph{Proceedings of The 27th International Conference on Artificial Intelligence and Statistics}, volume 238 of \emph{Proceedings of Machine Learning Research}, pages 208--216. PMLR, 02--04 May 2024{\natexlab{a}}.
\newblock URL \url{https://proceedings.mlr.press/v238/song24a.html}.

\bibitem[Song et~al.(2024{\natexlab{b}})Song, Zhang, and Zhang]{szz24}
Zhao Song, Lichen Zhang, and Ruizhe Zhang.
\newblock Training multi-layer over-parametrized neural network in subquadratic time.
\newblock In \emph{Innovations in Theoretical Computer Science (ITCS)}, 2024{\natexlab{b}}.
\newblock URL \url{https://arxiv.org/abs/2112.07628}.

\bibitem[Thirunavukarasu et~al.(2023)Thirunavukarasu, Ting, Elangovan, Gutierrez, Tan, and Ting]{thirunavukarasu2023large}
Arun~James Thirunavukarasu, Darren Shu~Jeng Ting, Kabilan Elangovan, Laura Gutierrez, Ting~Fang Tan, and Daniel Shu~Wei Ting.
\newblock Large language models in medicine.
\newblock \emph{Nature medicine}, 29\penalty0 (8):\penalty0 1930--1940, 2023.

\bibitem[Vaswani et~al.(2017)Vaswani, Shazeer, Parmar, Uszkoreit, Jones, Gomez, Kaiser, and Polosukhin]{vaswani2017attention}
Ashish Vaswani, Noam Shazeer, Niki Parmar, Jakob Uszkoreit, Llion Jones, Aidan~N Gomez, {\L}ukasz Kaiser, and Illia Polosukhin.
\newblock Attention is all you need.
\newblock \emph{Advances in neural information processing systems (NeurIPS)}, 30, 2017.
\newblock URL \url{https://arxiv.org/abs/1706.03762}.

\bibitem[Widrich et~al.(2020)Widrich, Sch{\"a}fl, Pavlovi{\'c}, Ramsauer, Gruber, Holzleitner, Brandstetter, Sandve, Greiff, Hochreiter, et~al.]{widrich2020modern}
Michael Widrich, Bernhard Sch{\"a}fl, Milena Pavlovi{\'c}, Hubert Ramsauer, Lukas Gruber, Markus Holzleitner, Johannes Brandstetter, Geir~Kjetil Sandve, Victor Greiff, Sepp Hochreiter, et~al.
\newblock Modern hopfield networks and attention for immune repertoire classification.
\newblock \emph{Advances in Neural Information Processing Systems (NeurIPS)}, 33:\penalty0 18832--18845, 2020.
\newblock URL \url{https://arxiv.org/abs/2007.13505}.

\bibitem[Williams(2018)]{williams2018some}
Virginia~Vassilevska Williams.
\newblock On some fine-grained questions in algorithms and complexity.
\newblock In \emph{Proceedings of the international congress of mathematicians: Rio de janeiro 2018}, pages 3447--3487. World Scientific, 2018.

\bibitem[Wu et~al.(2024{\natexlab{a}})Wu, Hu, Hsiao, and Liu]{wu2024uniform}
Dennis Wu, Jerry Yao-Chieh Hu, Teng-Yun Hsiao, and Han Liu.
\newblock Uniform memory retrieval with larger capacity for modern hopfield models.
\newblock In \emph{Forty-first International Conference on Machine Learning (ICML)}, 2024{\natexlab{a}}.
\newblock URL \url{https://arxiv.org/abs/2404.03827}.

\bibitem[Wu et~al.(2024{\natexlab{b}})Wu, Hu, Li, Chen, and Liu]{wu2023stanhop}
Dennis Wu, Jerry Yao-Chieh Hu, Weijian Li, Bo-Yu Chen, and Han Liu.
\newblock Stanhop: Sparse tandem hopfield model for memory-enhanced time series prediction.
\newblock In \emph{The Twelfth International Conference on Learning Representations (ICLR)}, 2024{\natexlab{b}}.
\newblock URL \url{https://arxiv.org/abs/2312.17346}.

\bibitem[Wu et~al.(2023)Wu, Irsoy, Lu, Dabravolski, Dredze, Gehrmann, Kambadur, Rosenberg, and Mann]{wu2023bloomberggpt}
Shijie Wu, Ozan Irsoy, Steven Lu, Vadim Dabravolski, Mark Dredze, Sebastian Gehrmann, Prabhanjan Kambadur, David Rosenberg, and Gideon Mann.
\newblock Bloomberggpt: A large language model for finance.
\newblock \emph{arXiv preprint arXiv:2303.17564}, 2023.
\newblock URL \url{https://arxiv.org/abs/2303.17564}.

\bibitem[Xu et~al.(2024)Xu, Huang, Hu, Li, Gilani, Goan, and Liu]{xu2024bishop}
Chenwei Xu, Yu-Chao Huang, Jerry Yao-Chieh Hu, Weijian Li, Ammar Gilani, Hsi-Sheng Goan, and Han Liu.
\newblock Bishop: Bi-directional cellular learning for tabular data with generalized sparse modern hopfield model.
\newblock In \emph{Forty-first International Conference on Machine Learning (ICML)}, 2024.
\newblock URL \url{https://arxiv.org/abs/2404.03830}.

\bibitem[Zhou et~al.(2023)Zhou, Ji, Li, Dutta, Davuluri, and Liu]{zhou2023dnabert}
Zhihan Zhou, Yanrong Ji, Weijian Li, Pratik Dutta, Ramana Davuluri, and Han Liu.
\newblock Dnabert-2: Efficient foundation model and benchmark for multi-species genome.
\newblock \emph{arXiv preprint arXiv:2306.15006}, 2023.
\newblock URL \url{https://arxiv.org/abs/2306.15006}.

\bibitem[Zhou et~al.(2024)Zhou, Wu, Ho, Wang, Shi, Davuluri, Wang, and Liu]{zhou2024dnabert}
Zhihan Zhou, Weimin Wu, Harrison Ho, Jiayi Wang, Lizhen Shi, Ramana~V Davuluri, Zhong Wang, and Han Liu.
\newblock Dnabert-s: Learning species-aware dna embedding with genome foundation models.
\newblock \emph{ArXiv}, 2024.
\newblock URL \url{https://arxiv.org/abs/2402.08777}.

\end{thebibliography}
\end{document}